\documentclass[10pt,twocolumn]{article}
\usepackage{hmi-preprint}
\usepackage{amsmath}
\usepackage{url}

\usepackage{booktabs}
\usepackage{array}
\usepackage{tabularx}
\usepackage{subcaption}
\usepackage{enumitem}
\usepackage{tikz}
\usetikzlibrary{matrix}
\usepackage{nicefrac}
\usepackage{microtype}
\usepackage{marvosym}         
\usepackage{xcolor}           
\usepackage{tocloft}
\definecolor{cvprblue}{rgb}{0.21,0.49,0.74}
\hypersetup{
    colorlinks=true, 
    linkcolor=cvprblue, 
    citecolor=cvprblue, 
    urlcolor=cvprblue,  
    pdftitle={Hand-Object Interaction in the Age of Large Foundation Models: Reconstruction, Generation, and Embodied Transfer},
    pdfauthor={Weiquan Lin, Yu Deng, Shiyang Liu, Luping Xiao, Xu Tang, Junzhi Yu, Jiaolong Yang, Lei Zhang, and Xingyu Chen},
    pdfsubject={A survey of foundation-model priors for hand-object interaction},
    pdfkeywords={hand-object interaction, reconstruction, generation, foundation-model priors, embodied transfer},
}

\newcolumntype{Y}{>{\raggedright\arraybackslash}X}
\newcolumntype{Z}{>{\centering\arraybackslash}X}

\setlist{topsep=2pt,itemsep=1pt,parsep=0pt,partopsep=0pt}

\preprintlogos{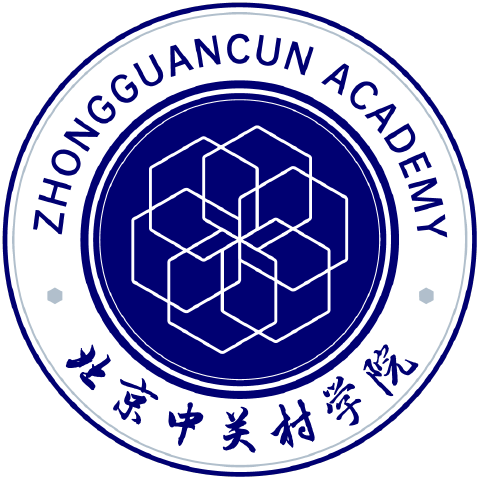}{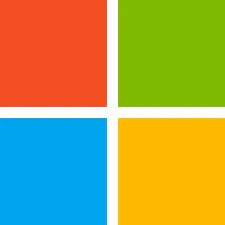}{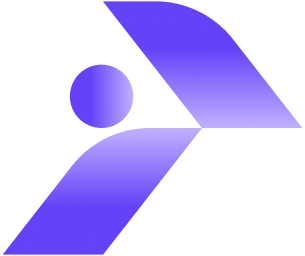}{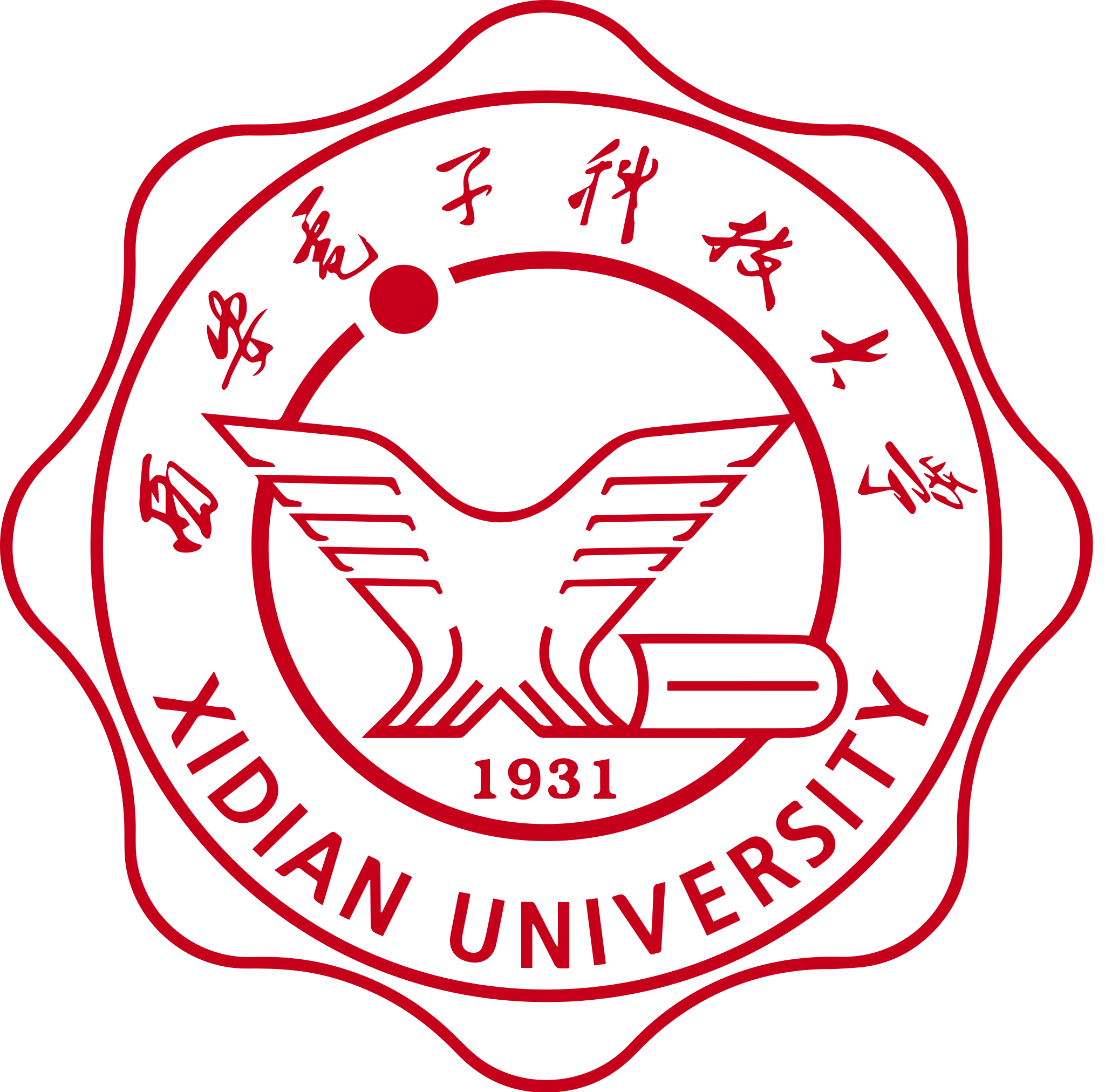}{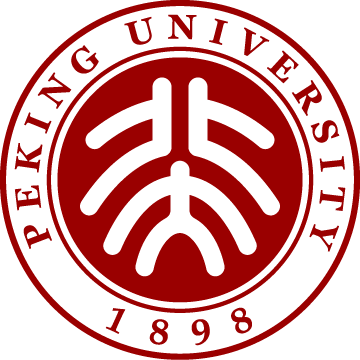}

\preprintlinkset{https://zgca-hmi-lab.github.io/HOISurvey}{https://github.com/SeanChenxy/Hand3DResearch/tree/hoi-survey}{}

\preprintdate{July 2026}

\runningtitle{Hand-Object~Interaction in the Age of Large Foundation Models: Reconstruction, Generation, and Embodied Transfer}

\preprintteaser{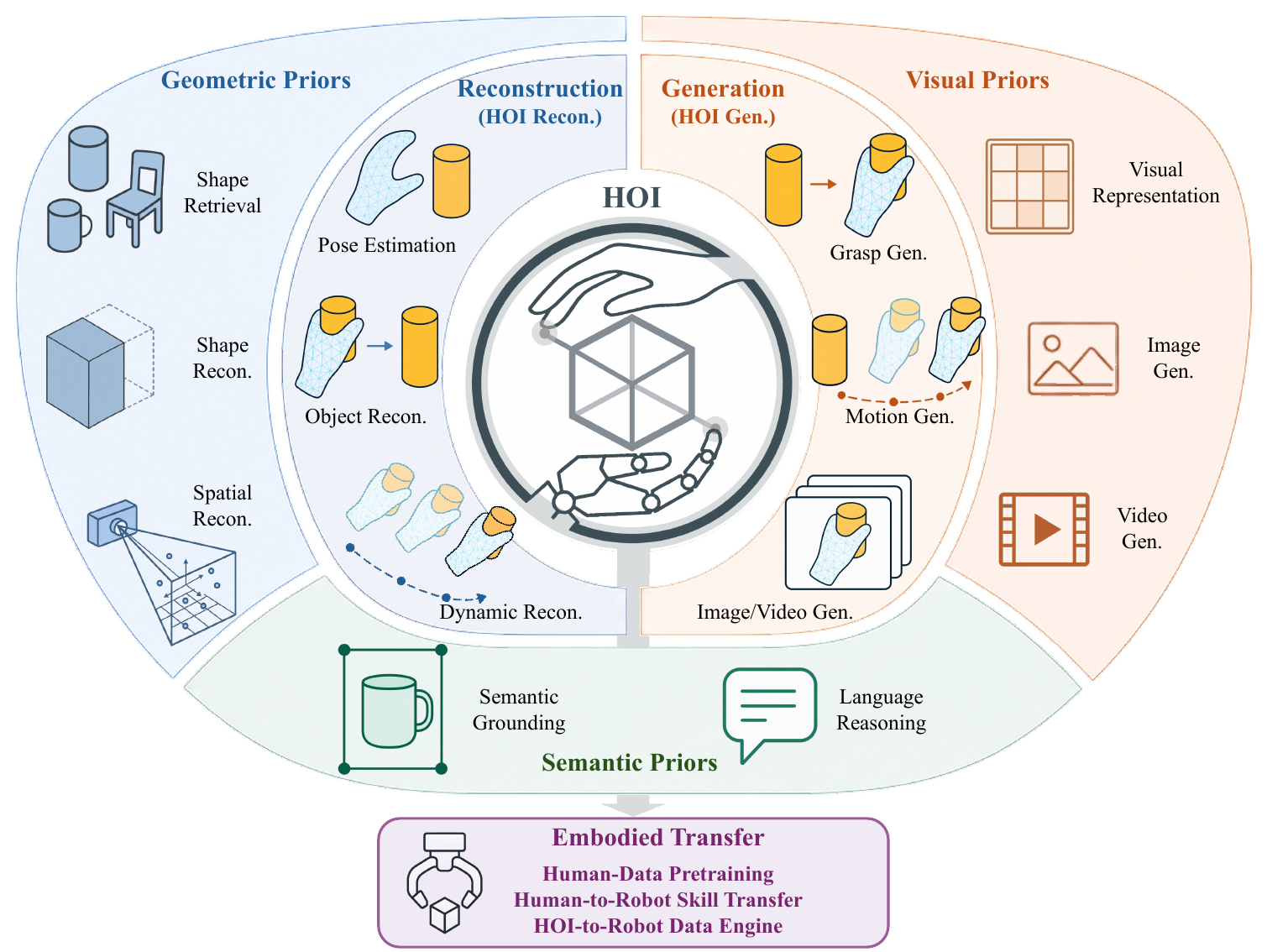}{Overview of this survey. The center organizes six HOI tasks into reconstruction (pose estimation, object reconstruction, and dynamic reconstruction) and generation (grasp, motion, and image/video generation). These tasks are supported by eight foundation-model sub-priors grouped into three families. Geometric priors cover shape retrieval, shape reconstruction, and spatial reconstruction. Semantic priors cover semantic grounding and language reasoning. Visual priors cover visual representation, image generation, and video generation. The resulting HOI knowledge further enables embodied transfer through human-data pretraining, human-to-robot skill transfer, and HOI-to-robot data generation.}

\title{Hand-Object~Interaction in the Age of Large Foundation Models: \\
Reconstruction, Generation, and Embodied Transfer
}

\author{
Weiquan Lin\textsuperscript{1,2} \hspace{0.15in}
Yu Deng\textsuperscript{3} \hspace{0.15in}
Shiyang Liu\textsuperscript{2} \hspace{0.15in}
Luping Xiao\textsuperscript{2} \hspace{0.15in}
Xu Tang\textsuperscript{1} \\
Junzhi Yu\textsuperscript{4} \hspace{0.15in}
Jiaolong Yang\textsuperscript{3} \hspace{0.15in}
Lei Zhang\textsuperscript{5} \hspace{0.15in}
Xingyu Chen\textsuperscript{2}
}

\affil{
\textsuperscript{1}Xidian University \hspace{0.12in}
\textsuperscript{2}Zhongguancun Academy \hspace{0.12in}
\textsuperscript{3}Microsoft Research Asia \hspace{0.12in} 
\textsuperscript{4}Peking University \hspace{0.12in}
\textsuperscript{5}Visincept \hspace{0.12in}
}

\begin{document}

\maketitle

\clearpage
\onecolumn
\thispagestyle{fancy}
\begin{center}
\tableofcontents
\end{center}
\clearpage
\twocolumn

\clearpage
\phantomsection
\addcontentsline{toc}{section}{Abstract}
\begin{abstract}
Hand-object interaction (HOI) modeling remains challenging because it requires joint reasoning about hand articulation, object geometry, contact, semantics, and dynamics under severe visual uncertainty.
Foundation models introduce transferable prior knowledge learned from large-scale cross-domain data, offering new ways to address these challenges beyond task-specific data and models. However, the rapidly growing literature remains fragmented, and existing studies typically describe these methods simply as ``using large models'' without systematically characterizing what knowledge is introduced, where it enters the HOI pipeline, or which HOI uncertainty it helps reduce.
This survey presents the first systematic review of foundation-model priors for HOI. We organize the literature into six HOI tasks spanning reconstruction and generation. 
More importantly, we establish a taxonomy of eight foundation-model sub-priors grouped into geometric, semantic, and visual families.
Geometric priors encompass shape retrieval, shape reconstruction, and spatial reconstruction; semantic priors include semantic grounding and language reasoning; and visual priors cover visual representation, image generation, and video generation.
Based on this taxonomy, we systematically analyze how different priors are represented, injected, and adapted across HOI pipelines and tasks. 
Beyond how foundation models empower HOI, we further examine how HOI-derived knowledge is used in robot learning, including human-data pretraining, human-to-robot skill transfer, and HOI-to-robot data generation.
Finally, we summarize datasets and evaluation protocols, and discuss limitations and future directions toward more generalizable HOI systems.
To support long-term progress, we curate a live repository that continuously aggregates emerging methods and benchmarks.
\end{abstract}

\noindent\textbf{Keywords:} Hand-Object Interaction, HOI Reconstruction, HOI Generation, Foundation-Model Prior, Embodied Intelligence


\section{Introduction}\label{sec:intro}

Hand-object interaction (HOI) is central to understanding how humans manipulate the physical world and underpins computer vision, augmented/virtual reality, digital humans, and embodied AI~\cite{DBLP:conf/cvpr/TekinBP19,hasson2019learning,yang2021cpf,fan2024hold}. HOI reconstruction and generation should jointly model hand articulation, object geometry, contact, interaction intent, and temporal evolution. Severe mutual occlusion and the coupling of pose, shape, depth, motion, semantics, and physical constraints make these variables only partially observable. Consequently, a plausible HOI result should be visually, geometrically, semantically, temporally, and physically consistent~\cite{ye2022s,brahmbhatt2020contactpose,liu2024grounding,hasson2021towards}.

These challenges manifest as five coupled uncertainties as follows.
\emph{Shape uncertainty} arises when the hand occludes object regions, leaving complete geometry ambiguous even across multiple views~\cite{ye2022s,karunratanakul2020grasping}. \emph{Spatial uncertainty} follows from the entanglement of pose, scale, depth, and viewpoint in monocular projection, which permits multiple 3D explanations of the same image~\cite{DBLP:conf/cvpr/TekinBP19,hasson2019learning}. \emph{Physical uncertainty} concerns both contact ambiguity and physical validity, i.e., visual proximity alone cannot determine contact, forces, or stability~\cite{brahmbhatt2020contactpose,grady2021contactopt,christen2022d}. \emph{Semantic uncertainty} covers open-world object identity, functional parts, affordances, and task intent~\cite{liu2024grounding,kirillov2023segment,li2024semgrasp}. \emph{Dynamic uncertainty} arises when camera, hand, and object motion become entangled under rapid viewpoint changes and motion blur~\cite{hasson2021towards,fan2024hold}. Because these uncertainties are coupled, reducing one of them in isolation does not guarantee an HOI result that is jointly consistent in geometry, contact, semantics, and motion; reliable HOI reconstruction and generation therefore require complementary evidence that constrains multiple variables and their interactions.

Foundation models provide such complementary cross-domain evidence beyond task-specific supervision. Rather than replacing HOI-specific models, they supply geometric, semantic, and visual knowledge that can be introduced at different stages of the HOI pipeline. Geometric priors support asset retrieval, shape recovery, and camera-aware spatial reconstruction; semantic priors provide open-vocabulary grounding and language reasoning; and visual priors transfer reusable representations or generative appearance and motion distributions~\cite{wang2024dust3r,liu2023openshape,liu2024grounding,bai2023qwen,oquab2023dinov2,rombach2022high,yang2025cogvideox}. This perspective shifts the central question from whether foundation models help HOI to \emph{which knowledge is introduced, how it enters the HOI pipeline, and which coupled uncertainties it helps reduce}.

We adopt a deliberately narrow operational boundary. A method is classified as foundation-model-prior only when an explicitly identified, large-scale general-purpose pretrained model contributes cross-domain knowledge through predictions, representations, transferred parameters, adaptation, or distillation. Task-specific pretraining and HOI-domain resources such as 3D assets, contact maps, or implicit fields do not alone meet this definition; they count as injected foundation-model priors only when a foundation model retrieves, generates, aligns, filters, or scores them for the target task~\cite{deitke2023objaverse,DBLP:conf/icml/RadfordKHRGASAM21,wu2026reconstructing}. All remaining methods are grouped as non-foundation-prior methods without implying an absence of pretraining or domain knowledge.

As shown in \autoref{fig:teaser}, our taxonomy has two dimensions. The task dimension comprises three reconstruction tasks (i.e., hand-object pose estimation, hand-held object reconstruction, and dynamic HOI reconstruction) and three generation tasks (i.e., grasp synthesis, HOI motion generation, and HOI image/video generation). The prior dimension contains eight sub-priors grouped into geometric (i.e., shape retrieval, shape reconstruction, and spatial reconstruction), semantic (i.e., semantic grounding and language reasoning), and visual (i.e., visual representation, image generation, and video generation) families. We further analyze how these priors enter HOI systems through initialization, alignment, fusion, conditioning, adaptation, or score guidance. Beyond reconstruction and generation, we examine how HOI videos, reconstructed states, grasps, trajectories, affordances, and generated plans transfer to robot learning through human-data pretraining, human-to-robot skill transfer, and HOI-to-robot data engines~\cite{ye2025latent,yang2025egovla,qin2022dexmv,liu2026egoengine}.

\begin{figure*}[!t]
\centering
\includegraphics[width=0.85\textwidth]{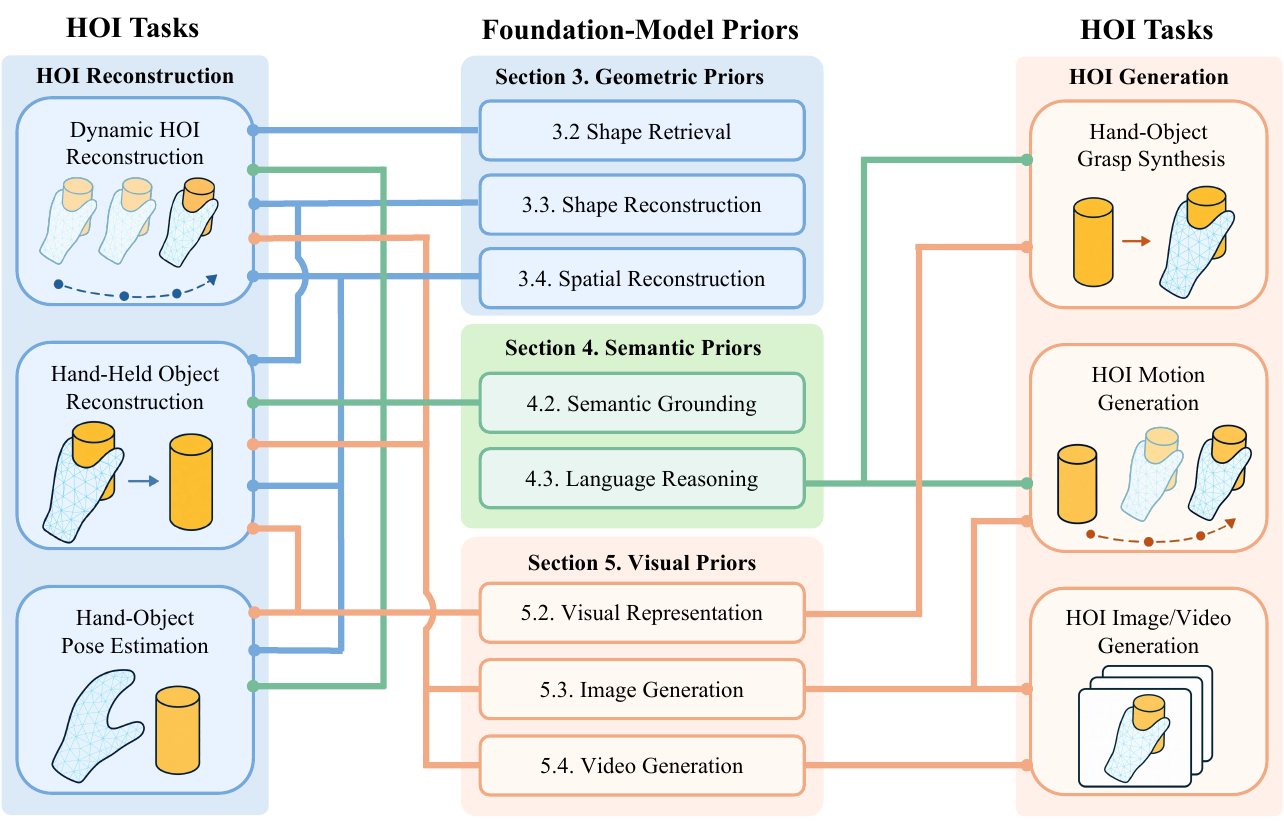}
\caption{Taxonomy roadmap of this survey. Three foundation-model prior families are decomposed into section-level sub-priors and linked to the HOI reconstruction or generation tasks reviewed under each subsection.}
\label{fig:priors-sections-tasks}
\end{figure*}

Existing surveys cover hand pose estimation, hand-object pose estimation, interacting hands, or dexterous grasping~\cite{ohkawa2023efficient,woo2023survey,miao2024advances,song2025overview}, but do not jointly analyze HOI reconstruction, generation, contact and affordance reasoning, foundation-model priors, and embodied transfer, as shown in \autoref{tab:survey-comparison}. This gap is increasingly consequential as recent pipelines combine heterogeneous foundation-model priors~\cite{liu2025easyhoi,wang2026arthoi}.

\begin{table*}[t]
\centering
\scriptsize
\setlength{\tabcolsep}{0.4pt}
\renewcommand{\arraystretch}{1.08}
\caption{Comparison of existing HOI-related surveys. $\checkmark$ = systematic coverage; $\times$ = no systematic coverage.}\label{tab:survey-comparison}
\begin{tabularx}{\textwidth}{@{}>{\raggedright\arraybackslash}X>{\centering\arraybackslash}p{0.059\textwidth}>{\centering\arraybackslash}p{0.069\textwidth}>{\centering\arraybackslash}p{0.060\textwidth}>{\centering\arraybackslash}p{0.070\textwidth}>{\centering\arraybackslash}p{0.074\textwidth}>{\centering\arraybackslash}p{0.059\textwidth}@{}}
\toprule
Existing survey & \shortstack{Hand-only\\pose} & \shortstack{Hand-object\\pose} & \shortstack{HOI\\generation} & \shortstack{Contact and\\affordance} & \shortstack{Foundation\\model priors} & \shortstack{Embodied\\transfer} \\
\midrule
Efficient annotation and learning for 3D hand pose estimation: A survey~\cite{ohkawa2023efficient} & $\checkmark$ & $\times$ & $\times$ & $\times$ & $\times$ & $\times$ \\
A Survey of Deep Learning Methods and Datasets for Hand Pose Estimation from Hand-Object Interaction Images~\cite{woo2023survey} & $\checkmark$ & $\times$ & $\times$ & $\times$ & $\times$ & $\times$ \\
Advances in Vision-Based Deep Learning Methods for Interacting Hands Reconstruction: A Survey~\cite{miao2024advances} & $\checkmark$ & $\times$ & $\times$ & $\times$ & $\times$ & $\times$ \\
An Overview of Learning-Based Dexterous Grasping: Recent Advances and Future Directions~\cite{song2025overview} & $\times$ & $\times$ & $\checkmark$ & $\checkmark$ & $\times$ & $\checkmark$ \\
\midrule
Ours & $\checkmark$ & $\checkmark$ & $\checkmark$ & $\checkmark$ & $\checkmark$ & $\checkmark$ \\
\bottomrule
\end{tabularx}
\end{table*}

To the best of our knowledge, this is the \textbf{first} systematic and critical survey of HOI techniques and their embodied transfer through the lens of foundation-model prior injection. The organization of this survey is illustrated in \autoref{fig:priors-sections-tasks}. The main contributions are as follows:
\begin{itemize}[leftmargin=*,topsep=2pt,itemsep=1pt]
\item \textbf{Tasks and Knowledge Boundary (\autoref{sec:tasks}).} We formalize a six-task HOI taxonomy and separate cross-domain foundation-model priors from task-specific pretraining and HOI-domain knowledge.
\item \textbf{Prior Taxonomy and Injection Mechanisms (Secs.~\ref{sec:3d-geometry-priors}--\ref{sec:generative-priors}).} We organize geometric, semantic, and visual priors and compare their sources, representations, injection mechanisms, supported tasks, mitigated uncertainties, and limitations.
\item \textbf{Embodied Transfer (\autoref{sec:robot-learning}).} We examine how HOI-derived knowledge supports human-data pretraining, human-to-robot skill transfer, and robot-data generation.
\item \textbf{Datasets and Evaluation (\autoref{sec:datasets}).} We consolidate task benchmarks, large-scale pretraining sources, and evaluation metrics while exposing current protocol blind spots.
\item \textbf{Open Challenges (\autoref{sec:challenges}).} We identify directions toward generalizable, interaction-aware, and physically grounded HOI and embodied manipulation.
\end{itemize}


\section{Preliminaries of Hand-Object Interaction and Foundation-Model Priors}\label{sec:tasks}

This section defines the task and representation conventions used throughout the survey, reviews methods that do not use cross-domain knowledge from foundation models, summarizes the foundation-model prior families employed across HOI tasks, and introduces a common pipeline abstraction for the six HOI tasks. These non-foundation-prior methods may use in-domain priors such as hand models (MANO~\cite{DBLP:journals/tog/0002TB17}, NIMBLE~\cite{li2022nimble}), object templates~\cite{DBLP:conf/cvpr/TekinBP19,hampali2020honnotate}, implicit fields~\cite{chen2022alignsdf,choi2024handnerf}, contact statistics~\cite{grady2021contactopt}, temporal smoothness~\cite{hasson2021towards,fan2024hold}, scene estimation (SLAM~\cite{durrant2006simultaneous}, SfM~\cite{snavely2006photo}), physical constraints or simulators~\cite{wang2023deepsimho,christen2022d}, and HOI generative models~\cite{karunratanakul2020grasping,li2025latenthoi}. Many of these components also appear inside foundation-model-prior methods as representations or refinement constraints. The distinction is whether a large-scale pretrained model supplies additional cross-domain knowledge.

\subsection{Task Taxonomy of Hand-Object Interaction}\label{sec:task-taxonomy}

This subsection defines the six HOI tasks and the terminology used throughout the survey. Contact and affordance are treated as output attributes rather than standalone task categories. For compactness in \autoref{tab:datasets} and \autoref{tab:metrics}, R1--R3 and G1--G3 denote the reconstruction and generation tasks defined as follows. R1 denotes Hand-Object Pose Estimation. R2 denotes Hand-Held Object Reconstruction. R3 denotes Dynamic HOI Reconstruction. G1 denotes Hand-Object Grasp Synthesis. G2 denotes HOI Motion Generation. G3 denotes HOI Image/Video Generation. ET denotes the downstream embodied-transfer setting discussed in \autoref{sec:robot-learning}.

\subsubsection{HOI Reconstruction}\label{sec:hoi-reconstruction-taxonomy}

HOI reconstruction recovers hand and/or object spatial state and structure from visual observations. We distinguish three tasks by output scope and temporal range:

\textbf{Hand-Object Pose Estimation (R1)} estimates the joint spatial state of the hand and the manipulated object, including hand articulation and object pose. Depending on the available observations and annotations, outputs range from hand keypoints, MANO parameters, and hand meshes to object 6D poses and joint hand-object poses. Hand-only methods are included because they often serve as front-end modules for HOI pipelines, where object fitting, contact estimation, and temporal optimization depend on reliable hand-state estimates~\cite{prakash20243d,DBLP:conf/cvpr/TekinBP19}. \textbf{Hand-Held Object Reconstruction (R2)} recovers the complete 3D geometry of an object held in the hand, using explicit representations such as point clouds or meshes, or implicit representations such as SDF, DDF, or occupancy fields. Its core challenge is inferring object regions occluded by the hand or otherwise unobserved, especially for unknown or open-world objects without a known CAD template~\cite{ye2022s,chen2022alignsdf,zhang2023ddf}. \textbf{Dynamic HOI Reconstruction (R3)} recovers time-varying hand, object, camera, contact, and interaction states from video. Renderable and 4D HOI reconstruction~\cite{fan2024hold,zhang2024ncrf,aytekin2026grasp} are treated as representation variants within this reconstruction task.

\subsubsection{HOI Generation}\label{sec:hoi-generation-taxonomy}

HOI generation creates hand-object states, motions, or visual content under specified conditions. We define three generation tasks by output modality, while treating editing, inpainting, object swapping, and reenactment as conditional generation rather than independent tasks.

\textbf{Hand-Object Grasp Synthesis (G1)} produces plausible static hand grasp poses, hand meshes, or contact configurations given objects, language, contact, or functional targets~\cite{corona2020ganhand,karunratanakul2020grasping,grady2021contactopt}. \textbf{HOI Motion Generation (G2)} generates manipulation sequences with time-varying hand motion, object motion, interaction phases, and state transitions~\cite{christen2022d,zhang2024artigrasp}. \textbf{HOI Image/Video Generation (G3)} synthesizes images or videos containing hand-object interactions, or performs conditional editing, inpainting, or reenactment under text, image, mask, pose, depth, or contact conditions~\cite{hu2022hand,zhang2024hoidiffusion}.

\subsection{Representations for Hand-Object Interaction}\label{sec:representations}

\subsubsection{Geometric Representations}\label{sec:geometric-reps}

Geometric representations describe the spatial state and shape of hands, objects, and scenes. Sparse representations such as 2D/3D keypoints and skeletons are efficient for pose estimation and tracking. Parametric hand models such as MANO~\cite{DBLP:journals/tog/0002TB17} and NIMBLE~\cite{li2022nimble} encode hand articulation and shape in a low-dimensional space, making them widely used in hand mesh regression and optimization. Explicit representations such as meshes and point clouds are common outputs for reconstruction. Implicit representations such as signed distance fields (SDFs), occupancy fields, and directed distance fields (DDFs)~\cite{park2019deepsdf,zhang2023ddf} model continuous surfaces. HALO~\cite{karunratanakul2021skeleton}, for example, maps a 3D hand skeleton to a differentiable articulated-hand occupancy field. Renderable representations such as NeRF~\cite{DBLP:conf/eccv/MildenhallSTBRN20} and 3D Gaussian Splatting~\cite{kerbl20233d} further support novel-view rendering and 4D HOI reconstruction.

\subsubsection{Interaction Representations}\label{sec:interaction-reps}

Interaction representations describe the relation between the hand, the object, and the task. Contact maps assign contact labels or probabilities to image pixels, point-cloud samples, or mesh vertices~\cite{yang2021cpf,DBLP:conf/eccv/TseZKLZC22}, whereas affordance maps identify object regions suitable for particular actions~\cite{jian2023affordpose}. Field-based representations define these relations continuously in 3D space: contact fields model contact or proximity at arbitrary query locations, while interaction fields encode broader spatial relationships between the hand and object~\cite{karunratanakul2020grasping,morales2025versatile}. Dense correspondences specify point-level hand-object associations~\cite{bansal2026hopformer}, while object-state transitions describe how an object moves or articulates during manipulation~\cite{liu2022hoi4d,fan2023arctic}.

\subsection{Non-Foundation-Prior Methods for HOI Reconstruction}\label{sec:non-fm-reconstruction}

\subsubsection{Hand-Object Pose Estimation}\label{sec:non-fm-ho-recon}

Hand-object pose estimation includes hand-only estimation and joint hand-object state recovery. Hand-only methods such as HandOccNet~\cite{park2022handoccnet}, MobRecon~\cite{chen2022mobrecon}, HaMeR~\cite{pavlakos2024reconstructing}, WiLoR~\cite{potamias2025wilor}, simpleHand~\cite{zhou2024simple}, and S\textsuperscript{2}HAND~\cite{chen2021model} estimate hand keypoints, MANO parameters, or meshes from a single RGB image, while KeypointFusion~\cite{liu2024keypoint} additionally fuses RGB and depth for 3D hand pose estimation. Joint methods such as H+O~\cite{DBLP:conf/cvpr/TekinBP19}, Hasson et al.~\cite{hasson2019learning}, HOPE-Net~\cite{doosti2020hope}, Keypoint Transformer~\cite{hampali2022keypoint}, THOR-Net~\cite{aboukhadra2023thor}, and HOISDF~\cite{qi2024hoisdf} combine MANO, known object geometry, keypoint reasoning, SDF constraints, differentiable rendering, or mutual exclusion. Contact-aware variants such as CPF~\cite{yang2021cpf}, S\textsuperscript{2}Contact~\cite{DBLP:conf/eccv/TseZKLZC22}, ContactArt~\cite{zhu2024contactart}, and CHOIR~\cite{morales2025versatile} additionally predict contact fields or maps, but still ultimately estimate or refine hand-object state.

The limitation is not that these methods lack structure, but that their estimates remain bounded by task supervision, MANO-like articulation, limited object coverage, weak contact labels, and dataset-specific contact statistics. Under monocular occlusion and open-ended manipulation goals, pose ambiguity, proximity-based contact errors, missing object geometry, and missing functional intent remain common failure modes. For joint hand-object estimation, shape retrieval and shape reconstruction priors address unknown-object geometry, spatial reconstruction priors address 3D pose and camera ambiguity, and semantic grounding and language reasoning priors address observation and intent ambiguity. These prior families are reviewed in Secs.~\ref{sec:shape-retrieval}--\ref{sec:spatial-reconstruction}, \autoref{sec:semantic-grounding}, and \autoref{sec:language-reasoning}.

\subsubsection{Hand-Held Object Reconstruction}\label{sec:non-fm-hho-recon}

Hand-held object reconstruction focuses on recovering the 3D shape of the manipulated object under severe hand occlusion. Representative methods include IHOI~\cite{ye2022s}, AlignSDF~\cite{chen2022alignsdf}, gSDF~\cite{chen2023gsdf}, DDF-HO~\cite{zhang2023ddf}, CHORD~\cite{li2023chord}, HandNeRF~\cite{choi2024handnerf}, HHOR~\cite{huang2022reconstructing}, HOLD~\cite{fan2024hold}, MOHO~\cite{zhang2024moho}, and TexHOI~\cite{aggarwal2025texhoi}. HandNeRF reconstructs a hand-object scene from a single RGB image, whereas the remaining methods span single- and multi-frame settings. These methods typically represent object geometry with SDF, occupancy, or DDF fields. They optimize the geometry using silhouette consistency, differentiable rendering, or multi-frame observations, then enforce hand-object compatibility through mutual-exclusion and contact constraints. Some additionally learn category-level shape priors from HOI data.

These tools make reconstruction possible but not open-ended. Persistent occlusion can leave interaction-critical object regions unobserved, category-level priors cannot cover arbitrary open-world objects, and better surface metrics do not guarantee contact consistency or physical plausibility. These limitations motivate the shape retrieval, shape reconstruction, and semantic grounding priors reviewed in \autoref{sec:shape-retrieval}, \autoref{sec:shape-reconstruction}, and \autoref{sec:semantic-grounding}.

\subsubsection{Dynamic HOI Reconstruction}\label{sec:non-fm-hom-recon}

Dynamic HOI reconstruction extends reconstruction to video, where hand motion, object motion, contact evolution, and camera motion are coupled. Representative methods include hand-only temporal reconstruction such as SeqHAND~\cite{yang2020seqhand} and Dyn-HaMR~\cite{yu2025dyn}. Joint hand-object pose estimation includes HOMAN~\cite{hasson2021towards}, InteractionFusion~\cite{zhang2019interactionfusion}, HOLD~\cite{fan2024hold}, and Tian et al.'s interaction-aware 4D Gaussian splatting~\cite{DBLP:journals/corr/abs-2511-14540}. Renderable or physics-aware reconstruction includes NCRF~\cite{zhang2024ncrf} and DeepSimHO~\cite{wang2023deepsimho}.

These methods typically combine temporal regularization, geometric estimates, and physical constraints, including motion smoothness, MANO consistency, rigid-body constraints, SLAM~\cite{durrant2006simultaneous}/SfM~\cite{snavely2006photo} camera estimates, multi-frame photometric optimization, contact continuity, stable-grasp assumptions, and simulation. Egocentric camera motion, blur, and long-horizon drift weaken these components, while renderable trajectories can still penetrate, lose contact, or produce physically incorrect object responses. These limitations motivate the spatial reconstruction and video generation priors reviewed in \autoref{sec:spatial-reconstruction} and \autoref{sec:video-gen-priors}, together with the contact and physical evaluation discussed in \autoref{sec:metrics-contact}.

\subsection{Non-Foundation-Prior Methods for HOI Generation}\label{sec:non-fm-generation}

\subsubsection{Hand-Object Grasp Synthesis}\label{sec:non-fm-grasp-gen}

Hand-object grasp synthesis generates static hand poses or contact configurations for objects. Representative methods include GanHand~\cite{corona2020ganhand}, GrabNet~\cite{taheri2020grab}, GraspTTA~\cite{jiang2021hand}, ContactOpt~\cite{grady2021contactopt}, ContactGen~\cite{liu2023contactgen}, Contact2Grasp~\cite{DBLP:conf/ijcai/0004LZLHCY23}, G-HOP~\cite{ye2024g}, ClickDiff~\cite{li2024clickdiff}, and FastGrasp~\cite{wu2025fastgrasp}. G-HOP learns a joint spatial diffusion prior over hand and object fields and applies it to both interaction reconstruction and static grasp synthesis. Different grasp-synthesis methods use different HOI-specific components, including object-conditioned distributions over the MANO model's hand-pose and hand-shape parameters, contact maps, implicit interaction fields, penetration penalties, local geometric constraints, and force-closure-inspired objectives.

The main limitation is functional ambiguity: a stable-looking grasp may still ignore whether the object should be held, pressed, turned, passed, or opened. Contact coverage and penetration penalties also remain imperfect proxies for execution stability, motivating the language reasoning priors reviewed in \autoref{sec:language-reasoning} and the embodied validation discussed in \autoref{sec:robot-learning}.

\begin{table*}[!t]
\centering
\scriptsize
\setlength{\tabcolsep}{3pt}
\renewcommand{\arraystretch}{1.08}
\caption{Functional abstraction of typical HOI backbones and task heads across the six tasks.}
\label{tab:hoi-pipeline}
\begin{tabularx}{\textwidth}{@{}>{\raggedright\arraybackslash}p{0.24\textwidth}>{\raggedright\arraybackslash}p{0.28\textwidth}>{\raggedright\arraybackslash}p{0.30\textwidth}Y@{}}
\toprule
Task & Typical HOI backbone & Typical task head & Canonical output \\
\midrule
R1: Hand-Object Pose Estimation & CNN encoder~\cite{DBLP:conf/cvpr/TekinBP19}; ViT encoder~\cite{pavlakos2024reconstructing}; geometry-aware transformer~\cite{hampali2022keypoint} & Keypoint/object-pose regressors~\cite{DBLP:conf/cvpr/TekinBP19}; MANO/mesh regressor~\cite{hasson2019learning,pavlakos2024reconstructing} & Hand pose; object 6D pose \\
R2: Hand-Held Object Reconstruction & Image encoder or neural renderer~\cite{choi2024handnerf}; multi-frame reconstruction optimization~\cite{fan2024hold}; implicit-field model~\cite{chen2022alignsdf,zhang2023ddf} & SDF/DDF/occupancy decoder~\cite{chen2022alignsdf,zhang2023ddf,li2023chord}; mesh/radiance decoder~\cite{huang2022reconstructing,choi2024handnerf} & Held-object geometry \\
R3: Dynamic HOI Reconstruction & Temporal tracking~\cite{hasson2021towards}; 4D optimization~\cite{fan2024hold}; 4D neural renderer~\cite{zhang2024ncrf,DBLP:journals/corr/abs-2511-14540} & Trajectory/camera/object-state head~\cite{hasson2021towards,fan2024hold}; deformation/rendering head~\cite{zhang2024ncrf,DBLP:journals/corr/abs-2511-14540} & Dynamic hand-object state \\
G1: Hand-Object Grasp Synthesis & Object encoder/conditional VAE~\cite{taheri2020grab}; diffusion model~\cite{wu2025fastgrasp,ye2024g} & Grasp/MANO head~\cite{taheri2020grab}; contact/score head~\cite{jiang2021hand,grady2021contactopt} & Static grasp pose or contact map \\
G2: HOI Motion Generation & Temporal transformer~\cite{cha2024text2hoi,zhang2025bimart}; diffusion motion model~\cite{christen2022d,li2025latenthoi} & Hand-object trajectory/sequence decoder~\cite{christen2022d,cha2024text2hoi,zhang2025bimart} & HOI motion sequence \\
G3: HOI Image/Video Generation & Image/video U-Net~\cite{zhang2024hoidiffusion,pang2025manivideo}; DiT/video transformer~\cite{dang2025svimo} & Image/video decoder~\cite{zhang2024hoidiffusion,dang2025svimo}; rendering head~\cite{qu2025hogsa} & HOI image/video \\
\bottomrule
\end{tabularx}
\end{table*}

\subsubsection{HOI Motion Generation}\label{sec:non-fm-motion-gen}

HOI motion generation produces dynamic manipulation sequences from motion capture, GRAB/OakInk-style datasets, or task-specific demonstrations. D-Grasp~\cite{christen2022d}, ArtiGrasp~\cite{zhang2024artigrasp}, GEARS~\cite{zhou2024gears}, LatentHOI~\cite{li2025latenthoi}, Gaze-guided HOI Synthesis~\cite{tian2024gaze}, How Do I Do That?~\cite{prakash2025synthesizing}, and BimArt~\cite{zhang2025bimart} learn motion or trajectory distributions from HOI data, interaction codebooks, local geometry, gaze/contact constraints, reinforcement learning, or physics simulation rather than from external motion or video foundation models. GEARS synthesizes hand motion sequences conditioned on hand and object trajectories.

These methods provide important comparison points, but their learned distributions are tied to the scale and diversity of available HOI motion data. Long-horizon object transitions, executable physical contact, and task-conditioned generalization to novel objects or intents remain difficult, motivating the language reasoning and video generation priors reviewed in \autoref{sec:semantic-priors} and \autoref{sec:generative-priors}.

\subsubsection{HOI Image/Video Generation}\label{sec:non-fm-visual-gen}

Hand-object image and video generation remains comparatively sparse before foundation-scale generative models. HOGAN~\cite{hu2022hand} directly generates hand-object images, while HOGSA~\cite{qu2025hogsa} uses mesh-based 3D Gaussian Splatting to render pose- and viewpoint-diverse bimanual HOI images for data augmentation. Mature task-specific video generation is largely absent: earlier GAN-based models struggle to preserve hand topology, object identity, contact relationships, temporal consistency, and open-vocabulary controllability. This gap directly motivates image and video generation priors in \autoref{sec:generative-priors}.

\begin{figure}[!t]
\centering
\includegraphics[width=\columnwidth]{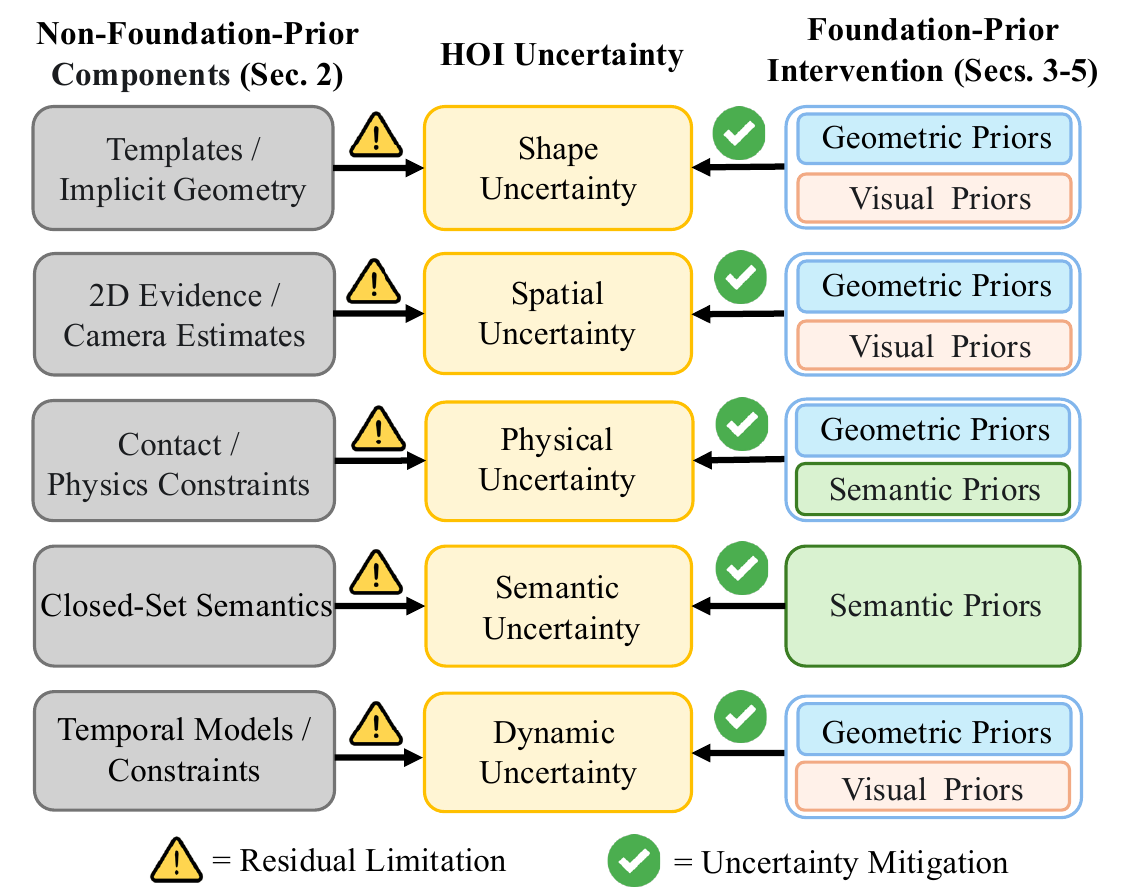}
\caption{Residual HOI uncertainties and corresponding foundation-prior interventions.}
\label{fig:traditional-foundation-comparison}
\end{figure}

\begin{table*}[!p]
\centering
\scriptsize
\setlength{\tabcolsep}{3pt}
\renewcommand{\arraystretch}{1.0}
\caption{Representative foundation-model-prior methods for HOI. The tag before P identifies the sub-prior defined in \autoref{sec:representations}; P and A denote primary and auxiliary priors, respectively. The representation and operator columns describe P. Uncertainty$\downarrow$ lists the HOI uncertainties mitigated by P: Sh = shape, Sp = spatial, Ph = physical, Se = semantic, and Dy = dynamic.}
\label{tab:representative-fm-methods}
\begin{tabularx}{\textwidth}{@{}>{\raggedright\arraybackslash}p{0.13\textwidth}>{\raggedright\arraybackslash}p{0.085\textwidth}>{\raggedright\arraybackslash}p{0.145\textwidth}>{\hsize=1.15\hsize\linewidth=\hsize\raggedright\arraybackslash}X>{\hsize=0.85\hsize\linewidth=\hsize\raggedright\arraybackslash}X>{\raggedright\arraybackslash}p{0.165\textwidth}>{\centering\arraybackslash}p{0.07\textwidth}@{}}
\toprule
Method & Status & Task & Foundation-model prior source & Injected representation & Injection operator & Unc.$\downarrow$ \\
\midrule
\multicolumn{7}{@{}l}{\textit{Primary family: Geometric priors} (\autoref{sec:3d-geometry-priors})} \\
\midrule
GeoHand~\cite{lin2026geohand} & ACMMM 2026 & R1: Hand-Object Pose Estimation & \textbf{G-Spa}; \textbf{P}: MoGe-2~\cite{wang2026moge} & spatial geometry tokens & token fusion & Sp \\
EasyHOI~\cite{liu2025easyhoi} & CVPR 2025 & R2: Hand-Held Object Reconstruction & \textbf{G-Rec}; \textbf{P}: InstantMesh~\cite{xu2024instantmesh}; \textbf{A}: LISA~\cite{lai2024lisa}, SAM~\cite{kirillov2023segment} (S-Gnd), Affordance Diffusion~\cite{ye2023affordance} (V-Img) & reconstructed shape & initialize & Sh \\
Follow My Hold~\cite{aytekin2026follow} & 3DV 2026 & R2: Hand-Held Object Reconstruction & \textbf{G-Rec}; \textbf{P}: Hunyuan3D~\cite{lai2025hunyuan3d}; \textbf{A}: MoGe-2~\cite{wang2026moge} & reconstructed shape & regularize & Sh \\
HOSt3R~\cite{swamy2025host3r} & ICCV Workshop 2025 & R2: Hand-Held Object Reconstruction & \textbf{G-Spa}; \textbf{P}: DUSt3R~\cite{wang2024dust3r} & pretrained parameters; point maps & weight initialization/fine-tuning & Sp \\
ArtHOI~\cite{wang2026arthoi} & CVPR 2026 & R3: Dynamic HOI Reconstruction & \textbf{G-Spa}; \textbf{P}: Video Depth Anything~\cite{chen2025video}, UniDepth~V2~\cite{piccinelli2025unidepthv2}; \textbf{A}: Hunyuan3D~\cite{lai2025hunyuan3d} (G-Rec), SAM~2~\cite{ravi2025sam}, Qwen-VL-Max~\cite{bai2023qwen} & metric depth / camera parameters & scale alignment / registration & Sp+Dy \\
GHOST~\cite{aboukhadra2026ghost} & CVPR 2026 & R3: Dynamic HOI Reconstruction & \textbf{G-Ret}; \textbf{P}: OpenShape~\cite{liu2023openshape}; \textbf{A}: SAM~2~\cite{ravi2025sam}, InternVL~\cite{chen2024internvl} & retrieved shape & retrieve / align / regularize & Sh \\
HaWoR~\cite{zhang2025hawor} & CVPR 2025 & R3: Dynamic HOI Reconstruction & \textbf{G-Spa}; \textbf{P}: Metric3D~\cite{yin2023metric3d} & metric depth & scale alignment & Sp+Dy \\
GraG~\cite{aytekin2026grasp} & arXiv 2026 & R3: Dynamic HOI Reconstruction & \textbf{G-Rec}; \textbf{P}: SAM~3D~\cite{chen2026sam}, MV-SAM3D~\cite{li2026mv}; \textbf{A}: SAM~3~\cite{carion2025sam}, Depth Anything~3~\cite{lin2025depth} & reconstructed shape & initialize / regularize & Sh+Dy \\
\midrule
\multicolumn{7}{@{}l}{\textit{Primary family: Semantic priors} (\autoref{sec:semantic-priors})} \\
\midrule
HandOS~\cite{chen2025handos} & CVPR 2025 & R1: Hand-Object Pose Estimation & \textbf{S-Gnd}; \textbf{P}: Grounding DINO~1.5~\cite{liu2024grounding} & region-aligned representation & region conditioning & Se+Sp \\
CHOIR~\cite{xu2026choir} & arXiv 2026 & R3: Dynamic HOI Reconstruction & \textbf{S-Gnd}; \textbf{P}: SAM~2~\cite{ravi2025sam}, Amodal Video Segmenter~\cite{chen2025using}; \textbf{A}: SAM~3D~\cite{chen2026sam} (G-Rec), MoGe-2~\cite{wang2026moge} (G-Spa) & modal / amodal masks & region conditioning & Se+Dy \\
Afford\-Grasp~\cite{wu2026affordgrasp} & CVPR 2026 & G1: Hand-Object Grasp Synthesis & \textbf{S-Lng}; \textbf{P}: RoBERTa~\cite{liu2019roberta}, Qwen2~\cite{yang2024qwen2technicalreport} & interaction semantics & condition / fuse & Se \\
NL2Contact~\cite{zhang2024nl2contact} & ECCV 2024 & G1: Hand-Object Grasp Synthesis & \textbf{S-Lng}; \textbf{P}: ChatGPT~\cite{chatgpt}, BERT~\cite{devlin2019bert} & interaction semantics & condition / fuse & Ph \\
HOIGPT~\cite{huang2025hoigpt} & CVPR 2025 & G2: HOI Motion Generation & \textbf{S-Lng}; \textbf{P}: LLaMA-13B~\cite{DBLP:journals/corr/abs-2302-13971} & interaction semantics & condition / fuse & Se \\
OpenHOI~\cite{zhang2026openhoi} & NeurIPS 2025 & G2: HOI Motion Generation & \textbf{S-Lng}; \textbf{P}: ShapeLLM~\cite{qi2024shapellm} & interaction semantics & condition / fuse & Se+Ph \\
\midrule
\multicolumn{7}{@{}l}{\textit{Primary family: Visual priors} (\autoref{sec:generative-priors})} \\
\midrule
HOPFormer~\cite{bansal2026hopformer} & ECCV 2026 & R1: Hand-Object Pose Estimation & \textbf{V-Rep}; \textbf{P}: DINOv2~\cite{oquab2023dinov2} & visual feature tokens & token fusion & Sp \\
ForeHOI~\cite{chen2026forehoi} & CVPR 2026 & R2: Hand-Held Object Reconstruction & \textbf{V-Rep}; \textbf{P}: DINOv2~\cite{oquab2023dinov2} & visual feature tokens & token fusion & Sh \\
MagicHOI~\cite{wang2025magichoi} & ICCV 2025 & R2: Hand-Held Object Reconstruction & \textbf{V-Img}; \textbf{P}: Zero-1-to-3~\cite{liu2023zero} & denoising score estimates & score-guided regularization & Sh \\
BIGS~\cite{on2025bigs} & CVPR 2025 & R3: Dynamic HOI Reconstruction & \textbf{V-Img}; \textbf{P}: Stable Diffusion~\cite{rombach2022high} & denoising score estimates; generative latent features & score-guided regularization; ControlNet conditioning & Sh+Dy \\
DreamHand~\cite{he2026dreamhand} & arXiv 2026 & R3: Dynamic HOI Reconstruction & \textbf{V-Vid}; \textbf{P}: Wan~\cite{wan2025wan} & generative latent features & weight initialization/fine-tuning & Dy \\
HUG~\cite{wu2026human} & arXiv 2026 & G1: Hand-Object Grasp Synthesis & \textbf{V-Rep}; \textbf{P}: DINOv2~\cite{oquab2023dinov2} & visual feature tokens & token fusion & Se+Sp \\
SIGHT~\cite{gavryushin2025sight} & arXiv 2025 & G2: HOI Motion Generation & \textbf{V-Img}; \textbf{P}: Stable Diffusion~\cite{rombach2022high}; \textbf{A}: CLIP~\cite{DBLP:conf/icml/RadfordKHRGASAM21} & denoising score estimates & score-guided guidance & Dy \\
SViMo~\cite{dang2025svimo} & NeurIPS 2025 & G3: HOI Image/Video Generation & \textbf{V-Vid}; \textbf{P}: CogVideoX-5B~\cite{yang2025cogvideox}; \textbf{A}: Google T5~\cite{raffel2020exploring} & pretrained parameters; generative latent features & weight initialization/fine-tuning; adapter conditioning & Dy \\
Re-HOLD~\cite{fan2025re} & CVPR 2025 & G3: HOI Image/Video Generation & \textbf{V-Img}; \textbf{P}: Stable Diffusion~\cite{rombach2022high}; \textbf{A}: LISA~\cite{lai2024lisa} & pretrained parameters; generative latent features & weight initialization/fine-tuning; ControlNet conditioning & Dy \\
Affordance Diffusion~\cite{ye2023affordance} & CVPR 2023 & G3: HOI Image/Video Generation & \textbf{V-Img}; \textbf{P}: GLIDE~\cite{nichol2022glide}, Latent Diffusion~\cite{rombach2022high} & pretrained parameters; generative latent features & weight initialization/fine-tuning & Se+Sh \\
\bottomrule
\end{tabularx}
\end{table*}

\subsection{Common HOI Pipeline Abstraction}\label{sec:common-hoi-pipeline}

The six HOI tasks share a functional pipeline: an input passes through an HOI backbone, and a task-specific head maps its representation to the required output. For generation, prompts, initial states, and other conditions are inputs. The HOI backbone denotes the main feature extraction, geometric inference, temporal modeling, optimization, or denoising process. The task head denotes the output-specific regressor, decoder, scorer, or renderer. \autoref{tab:hoi-pipeline} summarizes implementations across the six tasks.

The backbone and task head denote functional roles, not necessarily separate neural modules. Optimization-based reconstruction, implicit-field methods, and diffusion models may implement these roles through iterative optimization, neural fields, denoisers, or rendering decoders. Foundation-model priors preserve the HOI task definition while initializing, augmenting, conditioning, or instantiating the backbone or head. They may enter as observations, pretrained parameters, features, constraints, training data, or inference and decoding modules.

Conventional HOI components still leave five residual uncertainties under severe occlusion, open-world semantics, and complex dynamics. The three foundation-prior families reviewed in Secs.~\ref{sec:3d-geometry-priors}--\ref{sec:generative-priors} help mitigate these uncertainties through targeted interventions, as summarized in \autoref{fig:traditional-foundation-comparison}. Templates and implicit geometry leave occluded or unseen object surfaces ambiguous, whereas 2D evidence and camera estimates underconstrain depth, scale, camera motion, and world-space alignment. Geometric priors mitigate the resulting shape and spatial uncertainties. Visual priors provide additional evidence in these cases: generative denoising guidance can regularize weakly observed object geometry, while general visual representations can improve the estimation of hand-object pose and spatial state. Contact and physics constraints remain sensitive to inaccurate geometry and incorrectly inferred functional contact, so geometric and semantic priors jointly provide stronger structural and interaction evidence for reducing physical uncertainty. Closed-set supervision relies on predefined object categories and affordance labels, limiting its ability to generalize to unseen objects, functional parts, and interaction intents. Semantic priors mitigate this limitation through open-vocabulary grounding and language reasoning. Finally, task-specific temporal models and local consistency constraints, including motion smoothness, temporal pose consistency, cross-frame appearance consistency, and temporal contact consistency, remain insufficient for long-horizon interactions under moving cameras. Geometric priors stabilize camera-aware world-space structure, while visual priors contribute broader temporal regularities and interactive rollout models to mitigate dynamic uncertainty.


\subsection{Prior Family Employed in HOI Tasks}\label{sec:prior-families}

We summarize the three foundation-model prior families and their eight sub-priors used throughout the survey.

\subsubsection{Geometric Priors}\label{sec:geometric-priors-taxonomy}

Geometric priors transfer 3D shape and spatial structure from pretrained geometry models. \textbf{Geometric Shape Retrieval (G-Ret)} retrieves candidate object assets from shape databases to anchor object geometry~\cite{liu2023openshape}. \textbf{Geometric Shape Reconstruction (G-Rec)} generates complete object shapes from images or text~\cite{xu2024instantmesh,lai2025hunyuan3d}. \textbf{Geometric Spatial Reconstruction (G-Spa)} supplies metric depth, point maps, and camera parameters~\cite{wang2024dust3r,wang2026moge}.

\subsubsection{Semantic Priors}\label{sec:semantic-priors-taxonomy}

Semantic priors transfer visual and language understanding. \textbf{Semantic Grounding (S-Gnd)} converts text or multimodal prompts into localized regions such as masks, boxes, and tracks~\cite{liu2024grounding,kirillov2023segment}. \textbf{Language Reasoning (S-Lng)} infers intent, functional parts, and affordances from pretrained language models~\cite{achiam2023gpt,DBLP:journals/corr/abs-2302-13971}.

\subsubsection{Visual Priors}\label{sec:visual-priors-taxonomy}

Visual priors transfer cross-domain visual knowledge. \textbf{Visual Representation (V-Rep)} provides general-purpose image features or tokens~\cite{oquab2023dinov2}. \textbf{Image Generation (V-Img)} adapts pretrained diffusion backbones for appearance and layout~\cite{rombach2022high,nichol2022glide}. \textbf{Video Generation (V-Vid)} additionally models temporal and interaction dynamics~\cite{yang2025cogvideox,wan2025wan}.

\autoref{fig:traditional-foundation-comparison} summarizes the residual uncertainties and corresponding prior interventions, while \autoref{tab:representative-fm-methods} catalogs representative foundation-model-prior methods using the taxonomy developed in Secs.~\ref{sec:3d-geometry-priors}--\ref{sec:generative-priors}.


\section{Geometric Priors for HOI}\label{sec:3d-geometry-priors}

\begin{figure}[!t]
\centering
\includegraphics[width=0.88\columnwidth]{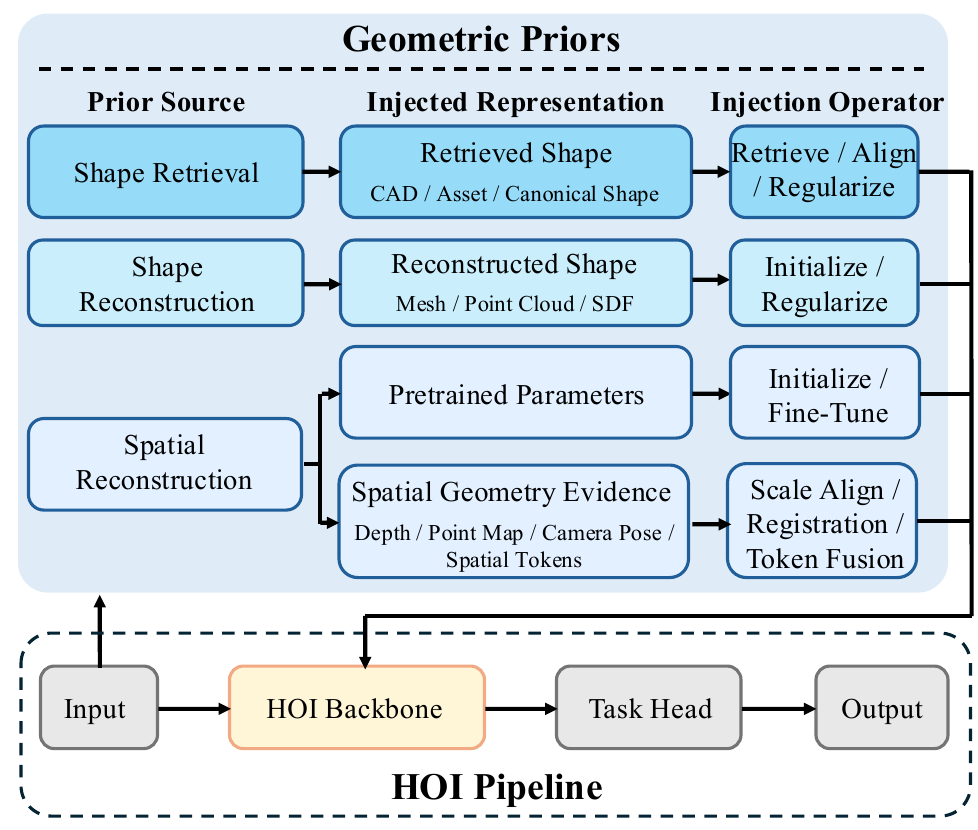}
\caption{Injection mechanisms of geometric priors for HOI.}
\label{fig:geometry-prior-injection}
\end{figure}

\subsection{Scope of Geometric Priors}\label{sec:3d-scope}

This section surveys foundation-model-derived 3D knowledge that reduces shape and spatial uncertainty in HOI. Three sub-priors are distinguished. Shape retrieval priors select topologically stable shape candidates from external 3D asset libraries by matching visual observations to assets using foundation-model embeddings. Shape reconstruction priors supply the occluded and unseen geometry of objects through single-image-to-3D or multi-view generative models that either complete partial evidence or synthesize an initial shape from scratch. Spatial reconstruction priors transfer pretrained parameters or provide depth, camera parameters, point maps, cross-view correspondences, and spatial feature tokens from pretrained spatial reconstruction models.

\autoref{fig:geometry-prior-injection} summarizes the corresponding injection mechanisms. Retrieved shapes are used through retrieval, alignment, and geometric regularization, while reconstructed shapes initialize or regularize the HOI solution. Spatial reconstruction priors either initialize and fine-tune model parameters or provide spatial geometry evidence for scale alignment, registration, and token fusion.

\subsection{Shape Retrieval Priors}\label{sec:shape-retrieval}

Shape retrieval priors select candidate object shapes from external 3D asset libraries by matching visual or language evidence to asset representations using foundation-model embeddings. In the methods surveyed here, InternVL~\cite{chen2024internvl} identifies the manipulated object, while OpenShape~\cite{liu2023openshape} retrieves its 3D hypothesis from Objaverse~\cite{deitke2023objaverse}.

\subsubsection{Dynamic HOI Reconstruction via Shape Retrieval}

Shape retrieval remains sparsely explored as the primary foundation-model prior for HOI. The representative example is GHOST~\cite{aboukhadra2026ghost}, in which InternVL~\cite{chen2024internvl} identifies the manipulated object and OpenShape~\cite{liu2023openshape} retrieves a corresponding asset from Objaverse~\cite{deitke2023objaverse}. The retrieved geometry is aligned with the video evidence and used to initialize and regularize the subsequent Gaussian reconstruction.

\subsubsection{Analysis and Discussion}
This route has not yet formed a broad method cluster in foundation-prior HOI. It requires a suitable asset to exist in the library, reliable object identification under hand occlusion, and robust geometric alignment across frames. These requirements are difficult to satisfy for instance-specific, articulated, deformable, or previously unseen objects. Consequently, retrieval more often serves as an auxiliary initialization or fallback hypothesis, whereas recent systems more commonly reconstruct or generate object geometry directly from the observations. When a suitable asset is available, retrieval provides a topologically stable object hypothesis that is easier to align and validate than a generated shape, though it may still differ from the target in geometry, scale, or articulation. Retrieval-based HOI reconstruction should therefore report retrieval failures separately from pose-fitting or contact-optimization errors and treat the retrieved asset as a confidence-weighted proposal rather than a guaranteed anchor.

\subsection{Shape Reconstruction Priors}\label{sec:shape-reconstruction}

Shape reconstruction priors leverage general 3D reconstruction and generation models to recover complete object geometry from partial observations or synthesize an initial shape for subsequent alignment. The methods surveyed here use InstantMesh~\cite{xu2024instantmesh}, Luma AI Genie~\cite{lumalabs}, Hunyuan3D~\cite{lai2025hunyuan3d}, SAM~3D~\cite{chen2026sam}, and MV-SAM3D~\cite{li2026mv}. Their reconstructed shapes provide hypotheses for occluded or unseen geometry but do not specify the true hand-object contact state and therefore require validation against the HOI observations.

\subsubsection{Hand-Held Object Reconstruction via Shape Reconstruction}\label{sec:shape-reconstruction-hho}

The primary application of shape reconstruction priors is in hand-held object reconstruction, where the hand occludes substantial portions of the object surface. Shape reconstruction methods use foundation-model-derived geometry to initialize or regularize the HOI backbone, helping it infer object surfaces that are missing from the visual observation.

EasyHOI~\cite{liu2025easyhoi} uses InstantMesh~\cite{xu2024instantmesh} to reconstruct a complete object mesh prepared by its upstream grounding and inpainting modules. This generated mesh initializes the subsequent hand-object alignment and physical optimization. MCC-HO~\cite{wu2026reconstructing} incorporates Retrieval-Augmented Reconstruction (RAR) as an internal component. RAR uses a GPT-4V~\cite{gpt4vsystemcard} object description to condition Luma AI Genie~\cite{lumalabs}, which generates a 3D object model. The resulting mesh initializes object reconstruction and is subsequently aligned across frames. Because the object hypothesis is generated rather than selected from an external asset library, MCC-HO is categorized primarily under shape reconstruction. Jiang et al.~\cite{jiang2025hand} use ChatGPT~\cite{chatgpt} to describe the held object and Luma AI Genie~\cite{lumalabs} to generate multiple textured mesh hypotheses. OpenShape~\cite{liu2023openshape} selects the generated candidate most consistent with the video, and DINOv2~\cite{oquab2023dinov2} features align its rendered views with each frame to initialize object poses before joint pose and implicit-shape optimization. Follow My Hold~\cite{aytekin2026follow} guides a Hunyuan3D~\cite{lai2025hunyuan3d} latent shape sample at inference time. HaMeR~\cite{pavlakos2024reconstructing} supplies the hand mesh, while MoGe-2~\cite{wang2026moge} supplies a partial point map and camera estimate. After these cues are registered, normal, disparity, silhouette, keypoint, intersection, and proximity losses jointly steer the diffusion velocity field and hand-object transforms.

\subsubsection{Dynamic HOI Reconstruction via Shape Reconstruction}\label{sec:shape-reconstruction-motion}

Shape reconstruction priors also serve video and 4D HOI reconstruction, where they initialize object geometry that is then tracked over time.

AGILE~\cite{shi2026agile} uses Hunyuan3D~\cite{lai2025hunyuan3d} to reconstruct an object mesh from selected multi-view observations, which initializes the object representation used by its temporal optimization. CHOIR~\cite{xu2026choir} uses SAM~3D~\cite{chen2026sam} to reconstruct a canonical object mesh with metric scale and an initial 6D pose before contact-aware 4D optimization. Grasp in Gaussians (GraG)~\cite{aytekin2026grasp} uses MV-SAM3D~\cite{li2026mv} to reconstruct canonical object geometry from selected keyframes, adapts SAM~3D~\cite{chen2026sam} to estimate per-frame object pose and scale while keeping the canonical shape fixed, and converts the dense Gaussian asset into a lightweight Sum-of-Gaussians representation for efficient tracking. In these pipelines, foundation-model-derived object shapes initialize the geometry used by subsequent pose estimation and tracking.

\subsubsection{Analysis and Discussion}
Shape reconstruction priors are most useful when hand occlusion hides interaction-critical object geometry. They expand the hypothesis space beyond category-specific HOI data, but reconstructed surfaces may hallucinate contact regions that conflict with hand-object pose, articulation, or physical feasibility. These priors are therefore best used for proposal generation or initialization, followed by pose fitting, contact and temporal verification, and physical plausibility filtering. Evaluation should complement Chamfer distance and F-score with contact correctness, penetration, and downstream interaction stability.

\subsection{Spatial Reconstruction Priors}\label{sec:spatial-reconstruction}

Spatial reconstruction priors recover 3D structure and camera geometry from input images or videos using pretrained spatial reconstruction models. They provide depth, camera parameters, point maps, cross-view correspondences, and camera trajectories, which can support the registration of hand-object states in a shared world coordinate system. The methods surveyed here use DUSt3R~\cite{wang2024dust3r}, VGGT~\cite{wang2025vggt}, CUT3R~\cite{wang2025continuous}, Metric3D~\cite{yin2023metric3d}, MoGe-2~\cite{wang2026moge}, Depth Anything~3~\cite{lin2025depth}, Video Depth Anything~\cite{chen2025video}, and UniDepth~V2~\cite{piccinelli2025unidepthv2}.

\subsubsection{Hand-Held Object Reconstruction via Spatial Reconstruction Priors}\label{sec:spatial-reconstruction-hho}

HOSt3R~\cite{swamy2025host3r} initializes a pairwise pointmap estimation network with DUSt3R~\cite{wang2024dust3r} pretrained weights and fine-tunes it on synthetic hand-object data. Its estimated pointmaps replace conventional SfM and keypoint matching when estimating hand-object 3D transformations and reconstructing the object from multiple views.

\subsubsection{Hand-Object Pose Estimation via Spatial Reconstruction}\label{sec:spatial-reconstruction-ho-recon}

GeoHand~\cite{lin2026geohand} freezes the foundational monocular geometry estimator MoGe-2~\cite{wang2026moge} and adapts it to 3D hand reconstruction through a GeoAdapter with gated cross-modal token fusion, showing that general spatial reconstruction knowledge can improve hand pose estimation under challenging monocular observations. HGGT~\cite{liu2026hggt} builds on VGGT~\cite{wang2025vggt} to jointly infer hand meshes and camera poses from uncalibrated images. Spatial priors do not recover complete hand-object states. However, their estimated hand geometry and camera parameters provide spatial cues that downstream HOI modules can use for object reconstruction, contact estimation, and motion recovery.

\subsubsection{Dynamic HOI Reconstruction via Spatial Reconstruction}\label{sec:spatial-reconstruction-motion}

Spatial reconstruction priors are particularly valuable for dynamic-camera settings, long-horizon sequences, and world-space reconstruction, where conventional tracking and SfM methods are prone to drift or failure.

HaWoR~\cite{zhang2025hawor} combines adaptive egocentric SLAM with Metric3D~\cite{yin2023metric3d}-based scale recovery for world-coordinate dynamic hand-only reconstruction. EgoGrasp~\cite{fu2026egograsp} combines Depth Anything~3~\cite{lin2025depth} depth and camera estimates with SAM~3D~\cite{chen2026sam} object geometry for world-space egocentric HOI estimation. ArtHOI~\cite{wang2026arthoi} injects metric depth and camera cues from Video Depth Anything~\cite{chen2025video} and UniDepth~V2~\cite{piccinelli2025unidepthv2}, together with Hunyuan3D~\cite{lai2025hunyuan3d} object geometry, into monocular 4D optimization of interactions between hands and articulated objects. In GraG~\cite{aytekin2026grasp}, Depth Anything~3~\cite{lin2025depth} supplies point maps and camera estimates that register the canonical object and its frame-wise motion. CHOIR~\cite{xu2026choir} uses MoGe-2~\cite{wang2026moge} metric depth to establish the scale of its SAM~3D-derived object initialization. Hand3R~\cite{hu2026hand3r} combines a frozen HaMeR hand expert with CUT3R~\cite{wang2025continuous}, using scene-aware visual prompting to fuse hand features with local scene tokens for online metric-scale hand-scene reconstruction from video.

\subsubsection{Analysis and Discussion}
Spatial reconstruction priors mainly address where the hand, object, camera, and scene lie in 3D space. Dense point maps, depth estimates, camera parameters, and cross-view correspondences can substantially reduce monocular scale-depth ambiguity, egocentric camera drift, and dynamic-view reconstruction errors. Their strength is not object semantics or interaction intent, but geometric stabilization: they provide a world-space scaffold on which HOI-specific modules can estimate contact, object motion, and interaction state. However, spatial reconstruction priors remain insufficient when the task requires knowing which object part should be used, whether apparent proximity corresponds to real contact, or whether a reconstructed motion is physically executable. They should therefore be viewed as sources of camera- and geometry-aware evidence for HOI pipelines, rather than as complete HOI reasoning systems. Future evaluations should report camera-pose errors, depth or point-map reconstruction quality, hand-object alignment, temporal drift, and downstream contact or object-state consistency.


\section{Semantic Priors for HOI}\label{sec:semantic-priors}

\subsection{Scope of Semantic Priors}\label{sec:semantic-scope}

This section focuses on identifying what the observed regions represent, which functional parts are involved, and what interaction intent they imply. Semantic priors inject foundation-model-derived visual and language understanding into HOI pipelines. We distinguish two complementary sub-priors by their function in the HOI pipeline. Semantic grounding priors convert open-vocabulary semantics, text prompts, or spatial queries into localized visual evidence, including boxes, masks, region-aligned representations, region correspondences, and temporal region tracks. Language reasoning priors transform natural language, functional knowledge, and interaction intent into constraints on grasp targets, contact regions, motion stages, and task semantics. The same VLM serves as a semantic grounding prior when used for region localization and as a language reasoning prior when used for interaction intent inference.

\begin{figure}[!t]
\centering
\includegraphics[width=0.88\columnwidth]{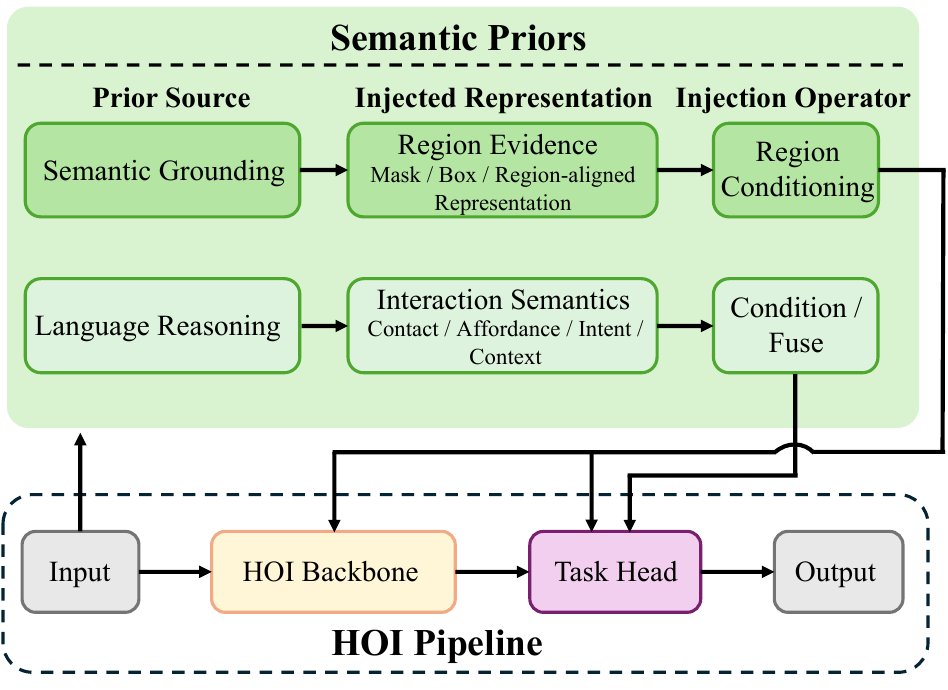}
\caption{Injection mechanisms of semantic priors for HOI.}
\label{fig:semantic-prior-injection}
\end{figure}

\autoref{fig:semantic-prior-injection} distinguishes two semantic intervention mechanisms: semantic grounding injects region evidence, including masks, boxes, and region-aligned representations, through region conditioning, whereas language reasoning injects affordance, intent, and contact cues as interaction constraints or scores.

\subsection{Semantic Grounding Priors}\label{sec:semantic-grounding}

The semantic grounding methods surveyed here use Grounding DINO~\cite{liu2024grounding} for text-prompted localization, SAM~\cite{kirillov2023segment}, SAM~2~\cite{ravi2025sam}, and SAM~3~\cite{carion2025sam} for promptable image and video segmentation, LISA~\cite{lai2024lisa} for reasoning-based segmentation, and the Amodal Video Segmenter~\cite{chen2025using} for recovering occluded object regions across video.

\subsubsection{Hand-Object Pose Estimation via Semantic Grounding}\label{sec:grounding-hand-pose}

Semantic grounding can serve directly as the reconstruction backbone for pose estimation rather than producing an external crop/mask. HandOS~\cite{chen2025handos} queries a frozen Grounding DINO~1.5~\cite{liu2024grounding} with the text prompt ``Hand'' and adapts its grounded detector features and instance queries through a hand-specific one-stage decoder to predict 2D joints, 3D hand vertices, and camera translation. ScaleHP~\cite{jing2026scalehp} likewise uses a frozen Grounding DINO with the generic prompt ``Hand'' to obtain multimodal features, localization queries, and reference points. Its metric 2D–3D decoder couples these outputs with 2D and 3D joint queries and a scale token, after which a perspective-constrained least-squares module recovers the global translation and metric hand pose in camera coordinates.

\subsubsection{Hand-Held Object Reconstruction via Semantic Grounding}\label{sec:grounding-hho}

EasyHOI~\cite{liu2025easyhoi} uses the segmentation models LISA~\cite{lai2024lisa} and SAM~\cite{kirillov2023segment} to produce hand and object masks that define the spatial scope for subsequent inpainting and 3D generation.

\subsubsection{Dynamic HOI Reconstruction via Semantic Grounding}\label{sec:grounding-motion}

In video and 4D HOI reconstruction, semantic grounding provides object localization, mask propagation, and dynamic region association. ArtHOI~\cite{wang2026arthoi} uses SAM~2~\cite{ravi2025sam} hand and object masks to localize the regions passed to its 4D optimizer. CHOIR~\cite{xu2026choir} combines SAM~2 with the diffusion-based Amodal Video Segmenter~\cite{chen2025using} to produce temporally complete object masks for contact-aware optimization. GraG~\cite{aytekin2026grasp} uses SAM~3~\cite{carion2025sam} concept grounding to extract hand and object masks from the input video. EgoGrasp~\cite{fu2026egograsp} and AGILE~\cite{shi2026agile} use SAM~2 masks to isolate the hand-object regions before applying their geometric reconstruction modules. GHOST~\cite{aboukhadra2026ghost} uses SAM~2 to propagate object masks across the interaction video before asset retrieval and Gaussian reconstruction.

\subsubsection{Analysis and Discussion}
Semantic grounding priors are most effective as high-recall observation extractors. They convert open-vocabulary text or multimodal prompts into masks, boxes, part regions, region tokens, or temporal tracks, thereby making hand-object pipelines less dependent on closed-set detectors and task-specific segmentation labels. This is particularly important for in-the-wild HOI, where the manipulated object, functional part, or interaction region may not belong to a predefined category. The main weakness is error propagation. A slightly inaccurate mask can mislead inpainting, shape generation, asset retrieval, contact estimation, and 4D optimization; errors are especially severe around fingertips, transparent objects, reflective surfaces, and heavily occluded contact regions. Semantic grounding should therefore be evaluated not only by detection or segmentation scores, but also by its downstream effect on reconstruction, contact, and generation quality.

\subsection{Language Reasoning Priors}\label{sec:language-reasoning}
This prior type contains two distinct semantic mechanisms. Reasoning-derived constraints use LLMs or MLLMs to infer functional parts, contact intent, task stages, or affordances. Language-aligned conditioning instead transfers reusable text and vision embeddings or tokens into a grasp or motion generator without requiring an explicit reasoning trace. Representative sources include GPT-4/4o~\cite{achiam2023gpt}, LLaMA~\cite{DBLP:journals/corr/abs-2302-13971}, Qwen~\cite{bai2023qwentech}, CLIP~\cite{DBLP:conf/icml/RadfordKHRGASAM21}, and multimodal models such as LLaVA~\cite{liu2023visual}, GPT-4V~\cite{gpt4vsystemcard}, and Qwen-VL~\cite{bai2023qwen}. In both cases, a pretrained cross-domain model is required to change the semantic condition, interaction constraint, or generated output; text labels or task-trained language encoders are insufficient.

\subsubsection{Hand-Object Grasp Synthesis via Language Reasoning}\label{sec:lang-grasp}

Language reasoning priors inject functional semantics into hand-object grasp synthesis, answering not only ``can this be grasped'' but also ``for what purpose.'' 

Text2Grasp~\cite{chang2025text2grasp} uses GPT-3~\cite{brown2020language} and CLIP~\cite{DBLP:conf/icml/RadfordKHRGASAM21} to convert task-level or personalized descriptions into part-level grasp plans. SemGrasp~\cite{li2024semgrasp} combines Vicuna~\cite{vicuna2023}, GPT-4/4V~\cite{achiam2023gpt,gpt4vsystemcard}, and CLIP~\cite{DBLP:conf/icml/RadfordKHRGASAM21} to map natural-language descriptions to discrete grasp representations. Multi-GraspLLM~\cite{li2024multi} aligns object point-cloud features with a Vicuna-based language backbone~\cite{vicuna2023} and autoregressively predicts grasp tokens for multiple robotic hands. NL2Contact~\cite{zhang2024nl2contact} converts ChatGPT-generated contact descriptions~\cite{chatgpt} into BERT embeddings~\cite{devlin2019bert} that condition hand-object contact-map generation. AffordGrasp~\cite{wu2026affordgrasp} combines RoBERTa~\cite{liu2019roberta} and Qwen2~\cite{yang2024qwen2technicalreport} conditioning with affordance-aware diffusion to constrain human grasp synthesis. G-DexGrasp~\cite{jian2025g} uses GPT-4o~\cite{openai2024gpt4o} to infer the affordance type and contact part specified by a task instruction, and GLIP~\cite{li2022glip} to localize that part on the object. The resulting contact-part evidence and retrieved pose-contact distributions guide MANO-based grasp generation and refinement for unseen object categories. AffordDexGrasp~\cite{wei2025afforddexgrasp} uses GPT-4o~\cite{openai2024gpt4o} to extract intention, object-part, and grasp-direction cues that condition open-set dexterous grasp synthesis.

\subsubsection{HOI Motion Generation via Language Reasoning}\label{sec:lang-motion}

Semantic priors also condition HOI motion generation, where language specifies manipulation goals, phase structure, interaction semantics, and the intended correspondence between text and motion. Here, explicit LLM/VLM reasoning and pretrained language-aligned conditioning are reported separately rather than treated as the same operation.

One group injects pretrained language-aligned embeddings into task-specific motion generators. Text2HOI~\cite{cha2024text2hoi} uses CLIP~\cite{DBLP:conf/icml/RadfordKHRGASAM21} text features to condition its 3D HOI motion generator, while an auxiliary contact module predicts hand-object surface-contact probabilities. DiffH2O~\cite{christen2024diffh2o} feeds a CLIP text embedding, together with an object-shape representation, to diffusion denoisers for grasping and manipulation motion. The semantic prior therefore enters through denoiser conditioning. JointHOI~\cite{song2026jointhoi} combines a CLIP text embedding with PointNet object features and converts them into prefix condition tokens for a Transformer denoiser that jointly generates bimanual motion, object motion, and dynamic contact maps. StructBiHOI~\cite{wang2026structbihoi} combines CLIP text embeddings with object features to condition its Mamba-based diffusion denoiser for long-horizon bimanual motion.

A second group adapts pretrained language models to represent interaction sequences or produce explicit intermediate constraints. HOIGPT~\cite{huang2025hoigpt} first encodes hand and object motion as factorized VQ-VAE tokens, then adapts LLaMA-13B~\cite{DBLP:journals/corr/abs-2302-13971} to autoregress over mixed text and HOI tokens. The language-model prior thus enters at the sequence-generation stage. OpenHOI~\cite{zhang2026openhoi} initializes its 3D multimodal LLM from ShapeLLM~\cite{qi2024shapellm} and fine-tunes it to predict spatial affordance maps and semantic task decompositions, which condition its downstream affordance-driven diffusion and physical refinement modules. SynHLMA~\cite{liu2025synhlma} adapts a pretrained Vicuna-7B model~\cite{vicuna2023} and aligns language embeddings with discrete manipulation tokens in a shared autoregressive sequence before decoding interactions with articulated objects. TOUCH~\cite{han2025touch} uses Qwen-7B~\cite{bai2023qwentech} to transform text into hierarchical coarse-to-fine conditions that enter its controllable HOI generator and explicit contact-map prediction branch. MEgoHand~\cite{zhou2026megohand} uses Eagle-2~\cite{li2025eagle2} to encode the task instruction and egocentric RGB context, and UniDepth~V2~\cite{piccinelli2025unidepthv2} to add metric spatial cues. Cross-modal attention fuses the resulting semantic and depth representations before a task-trained DiT flow-matching decoder generates fine-grained hand-object trajectories. Across these methods, general-purpose pretrained language models provide sequence-level guidance or intermediate affordance, task, and contact representations for HOI generation.

%

\subsubsection{Analysis and Discussion}
Language reasoning priors reduce semantic uncertainty by translating task descriptions, object affordances, functional knowledge, and intent into grasp targets, contact regions, motion stages, or interaction constraints. Their value lies in distinguishing visually similar but functionally different interactions, such as holding, passing, opening, pressing, or turning the same object. Nevertheless, semantic plausibility does not imply geometric or physical validity. A language model may correctly infer that a mug should be grasped by the handle, but it cannot by itself determine collision-free finger placement, stable force closure, frictional feasibility, or dynamically valid object motion. Language reasoning should therefore be grounded into spatial regions, contact candidates, trajectories, or simulator/robot validation loops. Future work should explicitly evaluate whether language priors improve functional contact and task success, rather than only improving text-output alignment or qualitative interpretability. ContactPrompt~\cite{jung2026training} and the affordance-guided diffusion prior of Suzuki et al.~\cite{suzuki2025affordance} further show that MLLM/VLM reasoning can produce contact-only or hand-only auxiliary signals for modular HOI systems.


\section{Visual Priors for HOI}\label{sec:generative-priors}

\subsection{Scope of Visual Priors}\label{sec:gen-scope}

This section surveys transferable visual knowledge from general-purpose pretrained models, where 
visual representation priors provide general image features or tokens, image generation priors model single-frame appearance and layout, and video generation priors additionally model temporal appearance, identity persistence, and interaction evolution.

\begin{figure}[!t]
\centering
\includegraphics[width=0.88\columnwidth]{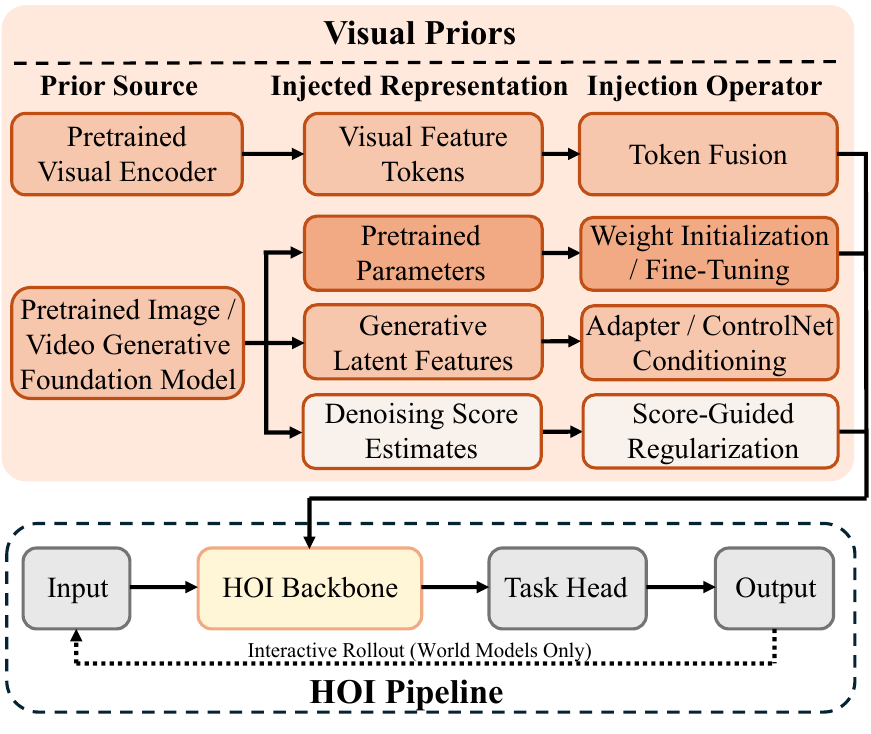}
\caption{Injection mechanisms of visual priors for HOI.}
\label{fig:visual-motion-prior-injection}
\end{figure}

\autoref{fig:visual-motion-prior-injection} summarizes three recurring injection paths for visual priors. General-purpose visual encoders provide feature tokens fused with HOI-specific representations; pretrained image and video generators contribute parameters or latent features through weight initialization, task-specific fine-tuning, and adapter- or ControlNet-based conditioning, and their denoising score estimates can regularize weakly observed HOI variables during optimization. In action-conditioned world models, predicted observations are additionally fed back as input for interactive rollout.

\subsection{Visual Representation Priors}\label{sec:visual-representation-priors}
Visual representation priors transfer general-purpose image features or tokens into an HOI model rather than producing masks, 3D geometry, or generated images. DINOv2~\cite{oquab2023dinov2} is the principal source in this group. Its visual features are combined with hand, object, or point-cloud representations through token fusion, which can be implemented using cross-attention.

\subsubsection{Hand-Object Pose Estimation via Visual Representation Priors}

HOPFormer~\cite{bansal2026hopformer} extracts object and scene tokens with DINOv2 ViT-G~\cite{oquab2023dinov2} and fuses them through cross-attention with pose-specialized WiLoR hand tokens~\cite{potamias2025wilor}. The fused representation supports joint estimation of two MANO hands and the 6D or articulated pose of the manipulated object. 

\subsubsection{Hand-Held Object Reconstruction via Visual Representation Priors}

HORT~\cite{chen2025hort} transfers DINOv2~\cite{oquab2023dinov2} visual tokens into hand-held object reconstruction, fusing them with hand geometry before predicting a coarse-to-fine object point cloud and its hand-relative pose. ForeHOI~\cite{chen2026forehoi} aggregates DINOv2 patch features across an interaction video and combines them with hand features to reconstruct the occluded object through feed-forward mask completion and 3D shape prediction. MCC-HO~\cite{wu2026reconstructing} uses DINOv2 features as an auxiliary visual prior for aligning its generated object geometry across frames. Jiang et al.~\cite{jiang2025hand} use DINOv2 features to align rendered views of the selected generated mesh with video frames during object-pose initialization.

\subsubsection{Hand-Object Grasp Synthesis via Visual Representation Priors}

Visual representation priors for grasp synthesis remain sparse. The representative example is Human Universal Grasping (HUG)~\cite{wu2026human}, which freezes DINOv2~\cite{oquab2023dinov2} image features, fuses them with metric point-cloud tokens, and generates open-world human grasps from RGB-D inputs and a query point. The resulting MANO grasps can subsequently be retargeted across robot hands.

\subsubsection{Analysis and Discussion}
The main value of this route is open-category robustness. DINOv2~\cite{oquab2023dinov2} features come from large-scale image pretraining, so they remain informative for object categories that HOI training sets rarely cover and degrade gracefully under hand occlusion. Across tasks, these features enter HOI models in two ways: they are fused through cross-attention with task-specific representations, such as pose-specialized hand tokens, hand geometry, and metric point-cloud tokens, or register generated and retrieved object geometry to video observations. However, appearance features carry no metric scale, articulation, or contact information, so they aid recognition and alignment but cannot resolve 3D structure or physical validity, and ambiguous appearance, such as textureless or transparent objects, can mislead the fusion stage. Evaluation should therefore emphasize unseen object categories rather than benchmark-only splits, together with the downstream effect on pose, reconstruction, and contact quality.

\subsection{Image Generation Priors}\label{sec:image-gen-priors}
The image generation methods surveyed here draw on GLIDE~\cite{nichol2022glide}, Stable Diffusion~\cite{rombach2022high}, SDXL~\cite{podell2024sdxl}, and Zero-1-to-3~\cite{liu2023zero} as pretrained diffusion backbones for image synthesis and novel-view generation. ControlNet~\cite{zhang2023adding} is also involved but used in a different role: it adds spatial conditioning to these backbones rather than serving as an independent foundation model.

Image generation also appears as an auxiliary prior in geometric-primary pipelines. EasyHOI~\cite{liu2025easyhoi} uses Affordance Diffusion~\cite{ye2023affordance} to reconstruct the object appearance hidden by the hand before InstantMesh produces the object mesh used by its geometric reconstruction stage.

\subsubsection{Hand-Held Object Reconstruction via Image Generation Priors}

MagicHOI~\cite{wang2025magichoi} uses a frozen Zero-1-to-3~\cite{liu2023zero} model to provide novel-view denoising guidance during joint hand-object reconstruction from short monocular videos. The resulting denoising score estimates regularize object geometry that is weakly observed from the input views. Diffusion-Guided Reconstruction~\cite{ye2023diffusion} initializes an image-conditioned diffusion backbone from GLIDE~\cite{nichol2022glide}, adapts it to hand-conditioned geometric renderings, and uses the resulting denoising score estimates to regularize complete object-shape optimization for each interaction clip. In both methods, the transferred prior constrains reconstruction through denoising score estimates rather than supplying a reconstructed shape from a 3D foundation model.

\subsubsection{Dynamic HOI Reconstruction via Image Generation Priors}

Image generation priors for dynamic HOI reconstruction remain sparse. The representative example is BIGS~\cite{on2025bigs}, which uses denoising score estimates from Stable Diffusion~\cite{rombach2022high} to regularize the hand-occluded regions of time-varying object Gaussians. ControlNet~\cite{zhang2023adding} supplies spatial conditioning during this optimization. This route must constrain weakly observed geometry while preserving object appearance and temporal consistency across the interaction sequence, which makes the direct use of image-generation priors in dynamic reconstruction uncommon.

\subsubsection{HOI Image/Video Generation via Image Generation Priors}\label{sec:image-gen-hoi}

HOIDiffusion~\cite{zhang2024hoidiffusion} adapts Stable Diffusion~\cite{rombach2022high} with CLIP~\cite{DBLP:conf/icml/RadfordKHRGASAM21}, ChatGPT~\cite{chatgpt}, and LLaVA~\cite{liu2023visual} conditions to generate 3D hand-object interaction data, demonstrating that image priors can synthesize training data for HOI reconstruction. Affordance Diffusion~\cite{ye2023affordance} initializes its LayoutNet from GLIDE~\cite{nichol2022glide} to sample plausible hand-object contact layouts, then fine-tunes a large-scale pretrained latent-diffusion inpainting model~\cite{rombach2022high} as ContentNet to render hand-object images conditioned on the object and sampled layout. Single-view to Novel-view Generation~\cite{zhang2025single} initializes its novel-view diffusion backbone from Zero123-XL~\cite{deitke2023objaversexl}, a scaled variant of Zero-1-to-3~\cite{liu2023zero} trained on Objaverse-XL, and injects hand depth and skeleton cues through a ControlNet-style conditioning branch~\cite{zhang2023adding}. Prompt-Propose-Verify~\cite{juneja2023prompt} uses GPT-4~\cite{achiam2023gpt} to enrich interaction prompts and DreamBooth-adapted SDXL proposers~\cite{ruiz2023dreambooth,podell2024sdxl} to construct an HOI image-text dataset, then fine-tunes SDXL for HOI image generation.

Beyond synthesizing complete interaction data, these priors also support hand-centric image generation and editing. Hand1000~\cite{zhang2025hand1000} fine-tunes Stable Diffusion~\cite{rombach2022high} with gesture-aware text embeddings for controllable hand-image generation. RHanDS~\cite{wang2025rhands} initializes its VAE and denoising U-Net from Stable Diffusion Inpainting v1.5~\cite{rombach2022high} and its structure encoder from a pretrained depth ControlNet~\cite{zhang2023adding} to repair malformed hands while preserving image style. AttentionHand~\cite{park2024attentionhand} adapts Stable Diffusion~\cite{rombach2022high} with text and rendered mesh conditions to generate training images for hand reconstruction.

\subsubsection{HOI Motion Generation via Image Generation Priors}
SIGHT~\cite{gavryushin2025sight} extends Stable Diffusion~\cite{rombach2022high} with CLIP~\cite{DBLP:conf/icml/RadfordKHRGASAM21} conditioning to generate 3D hand-object trajectories via inference-time diffusion guidance.

\subsubsection{Analysis and Discussion}
Image generation priors broaden the distributions of appearance, object category, background, and hand-object layout, but image fidelity does not ensure valid anatomy, contact, occlusion, or grasp function. Their value should therefore be assessed with interaction-aware criteria and downstream gains in pose estimation, contact prediction, grasp synthesis, or open-world generalization.

\subsection{Video Generation Priors}\label{sec:video-gen-priors}
The video generation methods surveyed here use DynamiCrafter~\cite{xing2024dynamicrafter}, CogVideoX~\cite{yang2025cogvideox}, and Wan~\cite{wan2025wan}, with FLUX.1~\cite{DBLP:journals/corr/abs-2506-15742} providing image-prior appearance initialization and inpainting conditions within these video pipelines, to transfer temporal appearance, object identity, and interaction dynamics into HOI tasks with or without interactive rollout.

\subsubsection{Dynamic HOI Reconstruction via Video Generation Priors}
DreamHand~\cite{he2026dreamhand} repurposes Wan~\cite{wan2025wan} to recover occlusion-robust 3D hand motion from egocentric videos, showing that video generation priors can support structured reconstruction when combined with geometric constraints.

\subsubsection{HOI Image/Video Generation via Video Generation Priors}\label{sec:video-gen-hoi}

Video generation priors have enabled a rapid expansion of HOI video generation, editing, and reenactment capabilities.

Re-HOLD~\cite{fan2025re} initializes its denoising and reference branches from Stable Diffusion v1.5~\cite{rombach2022high} and adds layout conditioning through a ControlNet-style branch~\cite{zhang2023adding} for HOI video reenactment. TASTE-Rob~\cite{zhao2025taste} fine-tunes DynamiCrafter~\cite{xing2024dynamicrafter} on task-oriented HOI videos to synthesize manipulation sequences with task-aligned hand-object dynamics. SViMo~\cite{dang2025svimo} adapts CogVideoX-5B~\cite{yang2025cogvideox} with T5 text features~\cite{raffel2020exploring} to jointly generate HOI videos and explicit 3D interaction sequences, feeding generated motion back into the diffusion process to improve video and motion consistency. iDiT-HOI~\cite{shen2025idit} combines a Wan-14B image-to-video prior~\cite{wan2025wan} with FLUX.1-dev~\cite{DBLP:journals/corr/abs-2506-15742} inpainting conditions for two-stage HOI reenactment. Yan et al.~\cite{yan2026open} adapt CogVideoX-I2V-5B~\cite{yang2025cogvideox} and Wan2.1-I2V-14B~\cite{wan2025wan} with structure- and contact-aware conditioning for open-world HOI video generation. HVG-3D~\cite{chen2026hvg} freezes a pretrained CogVideoX-5B-I2V backbone~\cite{yang2025cogvideox} and injects 3D point-cloud and tracking cues through a trainable 3D ControlNet with zero-initialized layers, enabling geometrically controlled HOI video synthesis. PAM~\cite{gao2026pam} initializes its appearance and motion stages from FLUX.1~\cite{DBLP:journals/corr/abs-2506-15742} and CogVideoX~\cite{yang2025cogvideox} and injects rendered depth, instance masks, and hand keypoints through ControlNet branches~\cite{zhang2023adding} for sim-to-real HOI video generation.

\subsubsection{From HOI Video Generation to HOI World Models}\label{sec:hoi-world-models}

A more recent line extends HOI video generation to action-conditioned interactive synthesis. Given the current observation and an action signal, these models predict future HOI frames without requiring the future object trajectory as an input. EgoHOI~\cite{li2026egocentric} fine-tunes Wan~2.1-14B~\cite{wan2025wan} with physics-informed 3D geometry and kinematic embeddings to predict egocentric HOI rollouts from a first frame and action signals. Hand2World~\cite{wang2026hand2world} adapts Wan2.1-1.3B-Control~\cite{wan2025wan} with a camera adapter and hand-gesture conditioning for autoregressive egocentric interaction rollout. Generated Reality~\cite{xie2026generated} conditions Wan2.2~\cite{wan2025wan} on joint hand and camera control to generate longer human-centric interactive videos. Dexterous World Models~\cite{kim2026dexterous} initializes from CogVideoX-Fun, an inpainting variant of CogVideoX~\cite{yang2025cogvideox}, and conditions the model on egocentric scene renderings and rendered hand-mesh motion. HandsOnWorld~\cite{chen2026handsonworld} fine-tunes Wan~\cite{wan2025wan} with protagonist-only 3D hand trajectories and a Pl\"ucker Hand Map, disentangling camera ego-motion from hand control in everyday interaction videos.

Wh0~\cite{chen2026wh0} fine-tunes Wan-I2V-A14B~\cite{wan2025wan} with language, object, and scene conditions to synthesize large-scale egocentric hand-object manipulation videos. It then extracts hand motion and applies visual editing to turn the generated videos into supervision for robot policy training.

\subsubsection{Analysis and Discussion}
Video generation priors model temporal appearance and identity across interaction phases, supporting augmentation, reenactment, planning, and interactive rollout. Beyond interactive synthesis, they can also act as scalable HOI data generators rather than direct interactive simulators, as in Wh0. This data-generation role differs from the generation in embodied transfer (\autoref{sec:robot-learning}), where HOI reconstruction or generation models produce task-aligned interaction outputs that directly serve robot learning; here, a general-purpose video prior supplies visual HOI data, and its conversion into robot supervision is a subsequent application. Temporal coherence, however, does not guarantee causal or physical validity: contact drift, implausible object motion, inconsistent articulation, and depth-ordering errors may persist. Evaluation should therefore complement video-quality metrics with contact persistence, object-state transition accuracy, occlusion consistency, and reconstruction or robot-execution performance.


\section{HOI-Derived Embodied Transfer}\label{sec:robot-learning}

\subsection{Scope of Embodied Transfer}\label{sec:robot-scope}

This section examines how HOI evidence and outputs from HOI reconstruction or generation support robot learning. Here, HOI evidence refers to interaction signals present in raw hand-object video and usable without explicit 3D recovery, such as hand-object motion, contact, and object-state changes. Outputs are the structured variables that HOI reconstruction or generation produces from this evidence, such as hand and object trajectories, contact maps, and generated interaction videos. Either form may directly supervise a robot policy, be adapted across embodiments, or be encoded as latent actions inferred from visual changes, where embodiment mapping aligns human-derived signals with robot morphology or action spaces. We include methods in which this knowledge supplies video-level supervision, explicit HOI variables, or interaction guidance that conditions robot decision-making.

\begin{figure}[!t]
\centering
\includegraphics[width=0.88\columnwidth]{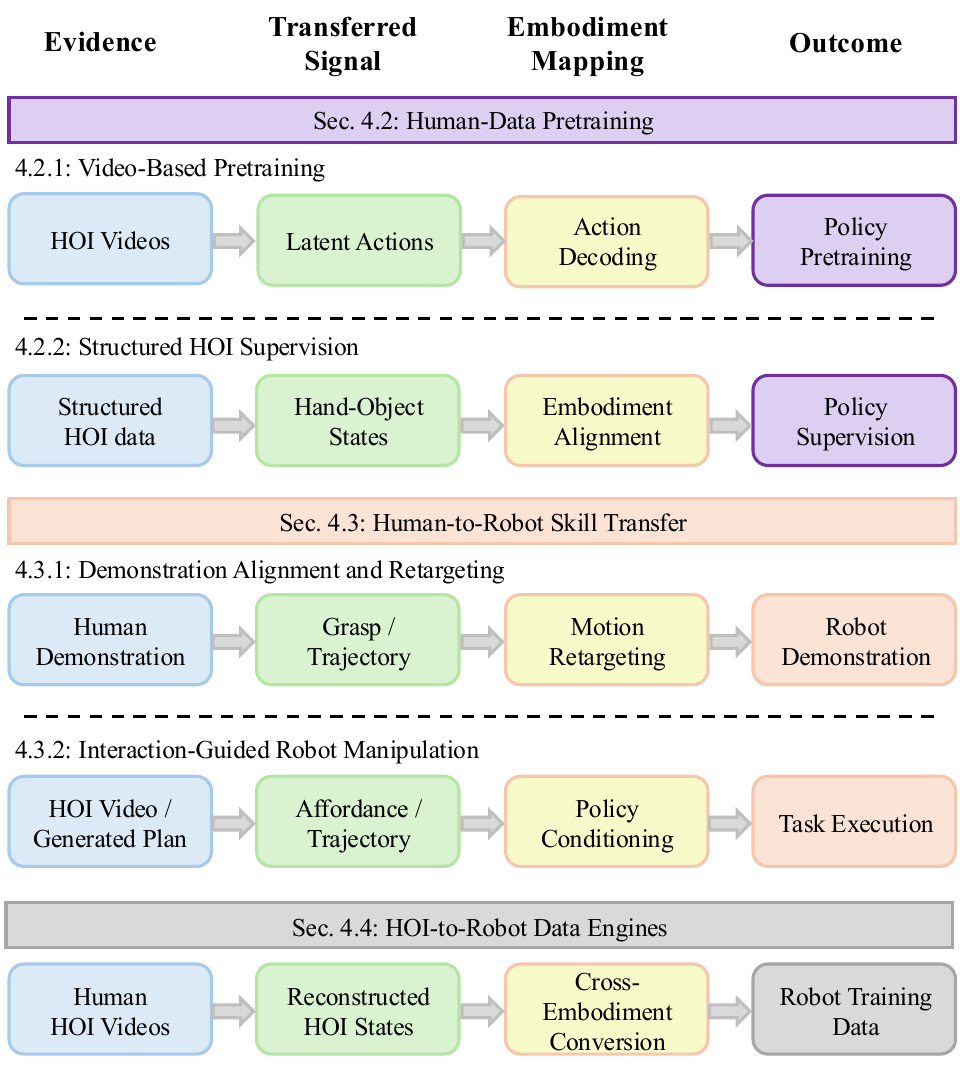}
\caption{Five routes by which HOI evidence becomes robot-learning supervision, target-skill guidance, or reusable robot training data.}
\label{fig:hoi-for-robot}
\end{figure}

\autoref{fig:hoi-for-robot} organizes five routes: video-based and structured human-data pretraining, demonstration retargeting and interaction-guided transfer for target skills, and HOI-to-robot data engines that produce reusable robot-aligned training data.

\subsection{Human-Data Pretraining}\label{sec:generalist}

\subsubsection{Video-Based Pretraining}\label{sec:video-pretraining}

Video-based pretraining is the least structured transfer route considered here. It does not require explicit reconstruction of hands, objects, or contact. Instead, manipulation-centric human videos provide visual state transitions from which a policy learns latent actions, image goals, or motion tokens. When clips visibly capture hand approach, contact, object response, and release, these representations provide weak HOI evidence alignable with robot actions. LAPA~\cite{ye2025latent} learns latent actions from unlabeled video transitions and decodes them into robot actions during downstream adaptation. Moto~\cite{chen2025moto} uses latent motion tokens as a bridge between video prediction and control. Video Prediction Policy~\cite{hu2024video} pretrains predictive visual representations on human manipulation videos through video prediction, then transfers these learned dynamics to generalist robot policies. IGOR~\cite{chen2024igor} studies image-goal representations, while In-N-On~\cite{cai2025n} and Emergence of Human to Robot Transfer~\cite{kareer2025emergence} study transfer from human video to robot policies.

When robot data, action vocabularies, or embodiment-specific adapters are available, human-video pretraining is explicitly aligned with robot action spaces. CLAP~\cite{zhang2026clap} aligns human-video transitions with a robot proprioceptive action codebook. UniVLA~\cite{bu2025univla} learns task-centric latent actions and adapts them across robot embodiments, and villa-X~\cite{chen2025villa} improves latent-action modeling and its integration into VLA pretraining. Large video-pretrained robot models provide additional evidence: GR-1~\cite{wu2024unleashing} and GR-2~\cite{cheang2024gr} transfer video-learned visual dynamics into policy fine-tuning; GR00T N1~\cite{bjorck2025gr00t} turns action-less manipulation video into policy supervision via latent actions or inverse dynamics; Motus~\cite{bi2026motus} derives pixel-level latent actions; and EgoScale~\cite{zheng2026egoscale} scales egocentric pretraining before human-robot mid-training.

Across these video-pretraining methods, scale improves coverage, but latent compression can discard fingertip contact, object 6D state, and causal physical effects. Their transfer should therefore be evaluated by robot-data efficiency and generalization to held-out tasks, objects, and embodiments.

\subsubsection{Structured HOI Supervision}\label{sec:structured-hoi}

Video pretraining usually learns implicit representations of visual and temporal changes rather than explicit hand-object variables. Structured HOI supervision makes these variables available as hand and wrist motion, 3D hand-object states, contact and affordance cues, frame-aligned action chunks, and 6DoF object trajectories, providing a direct interface between HOI analysis and human-data pretraining.

One group uses these variables directly as pretraining supervision. Being-H0~\cite{luo2025being} constructs part-level hand motion tokens with 3D physical alignment. VITRA~\cite{li2025scalable} converts human video into frame-aligned 3D hand motion and action chunks. EgoScaler~\cite{yoshida2025developing} extracts 6DoF object trajectories from egocentric videos and automatically refines noisy or incomplete trajectories, making interaction signals explicit.

Other methods combine structured HOI variables with robot-side alignment or policy training. EgoVLA~\cite{yang2025egovla} derives wrist and hand action targets from egocentric HOI video and maps them to robot actions via inverse kinematics and retargeting. H-RDT~\cite{bi2026h} uses 3D hand pose as privileged supervision before cross-embodiment fine-tuning. UniHM~\cite{zhang2026unihm} uses language-conditioned HOI states and actions for cross-hand and cross-task transfer. These mappings are more interpretable than latent video tokens.

\subsection{Human-to-Robot Skill Transfer}\label{sec:task-specific}

\subsubsection{Demonstration Alignment and Retargeting}\label{sec:dexterous-retargeting}

This route converts a human demonstration into a robot-aligned action sequence for a target skill. Unlike broad human-data pretraining, this route uses an intermediate HOI state that is reconstructed, inferred, or edited, such as a grasp, a hand-object trajectory, an object-state change, or a robotized visual observation, and explicitly aligns it across embodiments.

DexMV~\cite{qin2022dexmv} and Web2Grasp~\cite{chen2505web2grasp} use kinematic retargeting to convert inferred human demonstrations or functional grasps into dexterous-robot supervision. HUG~\cite{wu2026human} synthesizes MANO grasps from frozen DINOv2~\cite{oquab2023dinov2} features and metric point-cloud tokens and retargets them to different robot hands without per-embodiment training. GraspDreamer~\cite{tang2026graspdreamer} uses Veo~3.1, Kling Video~2.1, or Gemini~2.5 Image to synthesize functional grasp demonstrations and optimizes the recovered hand trajectories for robot execution. ManipTrans~\cite{li2025maniptrans} learns residual transfer from human bimanual motion to robotized demonstrations, whereas DexMachina~\cite{mandi2025dexmachina} preserves task function through object-state-centric retargeting. CHORD~\cite{zhu2026learning} instead aligns human and robot executions in object-centric contact-wrench space, using the induced force-torque effect as guidance for reinforcement learning. EgoMimic~\cite{kareer2025egomimic} jointly trains on human and robot demonstrations after 3D hand tracking, action normalization, and visual masking, avoiding a separate retargeting stage, while DexUMI~\cite{xu2025dexumi} maps wearable human-hand motion to robot hands. Object-Centric Dexterous Manipulation~\cite{chen2024object} couples a wrist-trajectory generator trained on human hand mocap with a reinforcement-learned finger controller to recover object-centric skills despite the embodiment gap. VideoManip~\cite{chen2026dexterous} reconstructs 3D hand-object trajectories from monocular RGB video and retargets them with contact optimization, training dexterous policies without wearable devices.

Instead of kinematic retargeting, some methods construct the demonstration itself from reconstructed or edited HOI states. DexImit~\cite{mu2026deximit} reconstructs bimanual demonstrations from monocular human video; Masquerade~\cite{lepert2025masquerade} edits human observations into robotized visual demonstrations; and YOTO~\cite{zhou2025you} extracts interaction states and keyframe trajectories from one human demonstration before expanding them into robot training data. Across these variants, skill-transfer quality should consider both cross-embodiment mapping fidelity and downstream task outcomes such as the intended object-state change; joint-angle similarity alone is insufficient.

\subsubsection{Interaction-Guided Robot Manipulation}\label{sec:interaction-guided}

Interaction-guided manipulation does not require frame-by-frame human-to-robot alignment. It transfers an embodiment-agnostic interaction description, such as an affordance, a contact target, a visual trajectory, or a generated manipulation video, that conditions robot decision-making.

Several methods condition decision-making on affordances or trajectories inferred from HOI video. GAT-Grasp~\cite{wang2025gat} infers affordances from HOI videos and transfers the resulting grasp targets to a robot grasping policy. ATM~\cite{DBLP:conf/rss/WenLS0D0A24} predicts future trajectories for arbitrary points in a video, then supplies those point trajectories as control guidance to a visuomotor policy, reducing the need for action-labeled demonstrations. VidBot~\cite{chen2025vidbot} recovers metric-scale 3D hand trajectories from in-the-wild videos via monocular depth and structure-from-motion, then learns language-conditioned 3D affordances that guide manipulation across embodiments. FlowHOI~\cite{zeng2026flowhoi} generates semantically aligned hand-object motion and uses the generated motion as guidance for dexterous manipulation.

Flow-based transfer methods make the intermediate object motion more explicit. Im2Flow2Act~\cite{xu2024flow} extracts object points from human demonstration videos, predicts their future trajectories, and uses the resulting object flow as the interface for generating robot actions. 3DFlowAction~\cite{zhi2025flowaction} learns a language-conditioned 3D-flow world model from human and robot manipulation data, predicts future object flow, and uses that prediction to constrain closed-loop robot action optimization. NovaFlow~\cite{li2025novaflow} and Dream2Flow~\cite{dharmarajan2025dream2flow} similarly distill generated videos into 3D object flow or motion, which is converted into grasp proposals, trajectory optimization, or low-level robot commands for rigid and deformable objects.

Point-track methods use a related interface without requiring a dense object-flow field. 3PoinTr~\cite{hung2026pointr} predicts embodiment-agnostic 3D point tracks from casual human videos, compresses them into point-track tokens, and trains a robot policy to follow these tokens with a small amount of robot action data. Dex4D~\cite{kuang2026dex4d} obtains object-centric point tracks from a human video or a generated point forecast, trains a point-to-action policy in simulation, and executes it with a real-time tracker for closed-loop sim-to-real manipulation.

A separate route uses generated visual interaction plans. Gen2Act~\cite{bharadhwaj2024gen2act} directly uses the pretrained VideoPoet model~\cite{kondratyuk2023videopoet} to generate a language-conditioned human manipulation video, then feeds that video to a robot policy as a visual task plan. TASTE-Rob~\cite{zhao2025taste} (cf.~\autoref{sec:video-gen-priors}) refines its generated hand-object motion and uses the resulting videos as demonstrations for imitation learning.

\subsection{HOI-to-Robot Data Engines}\label{sec:data-engine}

The methods above either pretrain policies from human HOI evidence or transfer that evidence into a single target skill. A complementary line converts real human HOI evidence into reusable cross-embodiment datasets, robotized videos, or executable demonstrations for training robot policies across different learning algorithms. Recent systems also study the data-acquisition side of embodied transfer. ACE-Data-0~\cite{cao2026acedata} introduces an ambient capture engine that synchronizes egocentric and exocentric video, body and hand motion, object geometry and 6-DoF trajectories, audio, and tactile signals in home environments, with a hierarchical benchmark connecting sensory signals to HOI.

Beyond data capture, other systems convert the recorded HOI evidence into robot-side training data. RoboWheel~\cite{zhang2026robowheel} converts monocular human HOI videos into robot data through trajectory reconstruction and cross-embodiment retargeting. EgoEngine~\cite{liu2026egoengine} produces dexterous robot demonstrations, while Ego2Robot~\cite{wang2026ego2robot} emphasizes scalable action and visual alignment across robot morphologies. HandEdit~\cite{yang2026handedit} benchmarks human-to-robot visual editing across 26 embodiments. RoboEdit~\cite{liu2026roboedit} turns human videos into robot experience through video editing. Do as I Do~\cite{paliwal2026dexterous} reconstructs hand-object trajectories and retargets them for robot manipulation. EgoInfinity~\cite{wang2026egoinfinity} scales to web-scale HOI evidence with any-view retargeting. Human2Robot~\cite{xie2026human2robot} translates human manipulation into robot-aligned visual demonstrations. TraceGen~\cite{lee2025tracegen} converts videos into 3D trace-space supervision. H2R-Grounder~\cite{ci2025h2r} translates human videos into physically grounded robot videos. Qwen-RobotManip~\cite{yuan2026qwen} retargets hand trajectories for cross-platform robot demonstrations.

These engines differ in the robot-side data they produce, including state-action trajectories, paired demonstrations, and robotized videos, but share the challenge of preserving task-relevant HOI state while bridging visual, kinematic, and physical embodiment gaps. Evaluation should separate HOI reconstruction fidelity, visual realism, action executability, cross-embodiment coverage, and downstream data efficiency.


\section{Datasets and Evaluation Protocols}\label{sec:datasets}

We survey benchmark datasets, pretraining sources, and evaluation protocols used across the HOI task taxonomy and embodied transfer, summarize the metric families commonly reported for each task and their blind spots, and reuse the abbreviations R1--R3, G1--G3, and ET introduced in \autoref{sec:task-taxonomy}.

\subsection{Datasets and Pretraining Sources}\label{sec:benchmark-datasets}

\autoref{tab:datasets} presents two parallel groups: task-level HOI benchmarks used to evaluate reconstruction and generation methods and large-scale human-centric data sources used in model development and embodied transfer.

For the large-scale sources, the ``Use'' column distinguishes video-model pretraining from human-data policy pretraining. The former learns general visual and temporal representations from human-centric videos, whereas the latter uses human interaction data to pretrain robot policies or action representations. These labels describe data use only; method-level classification follows \autoref{sec:robot-learning}.

\begin{table*}[!t]
\centering
\scriptsize
\setlength{\tabcolsep}{2.6pt}
\renewcommand{\arraystretch}{1.09}
\caption{Benchmark datasets and pretraining sources used in HOI research. Panel A summarizes released observations, camera view, acquisition setting, ground-truth (GT) annotations, and use; Panel B summarizes supervision and use for large-scale pretraining corpora. Observation denotes raw sensor streams or released input assets rather than 3D annotations. View --- ego: first-person; exo: external third-person; multi-view: synchronized camera rig; object: object-centric; N/A: no camera view. Setting --- controlled, in-the-wild, synthetic, web, or mixed. ``--'' indicates unavailable annotations.}
\label{tab:datasets}
\resizebox{\textwidth}{!}{%
\begin{tabular}{@{}lclllllll@{}}
\toprule
\multicolumn{9}{@{}l}{\textbf{A. Benchmark datasets}} \\
\midrule
Dataset & Year & Observation & View & Setting & Hand GT & Object GT & HOI GT & Task \\
\midrule
Dexter+Object~\cite{sridhar2016realtime} & 2016 & RGB-D & exo & controlled & fingertips & cuboid corners & -- & R1, R3 \\
EgoDexter~\cite{mueller2017realtime} & 2017 & RGB-D & ego & in-the-wild & fingertips & -- & -- & R1, R3 (hand-only) \\
FPHA~\cite{garcia2018first} & 2018 & RGB-D & ego & controlled & 3D joints & 6D pose (subset) & action & R1, R3, ET \\
FreiHAND~\cite{zimmermann2019freihand} & 2019 & RGB & exo & mixed & MANO & -- & -- & R1 (hand-only) \\
ObMan~\cite{hasson2019learning} & 2019 & RGB-D & exo & synthetic & MANO & shape+pose & contact & R1, R2, G1 \\
ContactDB~\cite{brahmbhatt2019contactdb} & 2019 & thermal & object & controlled & -- & object mesh & measured contact & G1 \\
HO3D~\cite{hampali2020honnotate,DBLP:journals/corr/abs-2107-00887} & 2020 & RGB & exo & controlled & MANO & 6D pose+CAD & -- & R1, R3 \\
InterHand2.6M~\cite{DBLP:conf/eccv/MoonYWSL20} & 2020 & RGB & exo & controlled & MANO (2H) & -- & -- & R1 (hand-only) \\
ContactPose~\cite{brahmbhatt2020contactpose} & 2020 & RGB-D & multi-view & controlled & MANO & 6D pose & contact & R1, G1 \\
GRAB~\cite{taheri2020grab} & 2020 & MoCap & N/A & controlled & SMPL-X & mesh+motion & contact/action & G1, G2 \\
MOW~\cite{cao2021reconstructing} & 2021 & RGB & exo & in-the-wild & 3D joints & shape+pose & pseudo 3D & R2 \\
DexYCB~\cite{chao2021dexycb} & 2021 & RGB-D & multi-view & controlled & MANO & 6D pose+CAD & -- & R1, R3 \\
H2O~\cite{kwon2021h2o} & 2021 & RGB-D & ego+multi-view & controlled & MANO (2H) & 6D pose+mesh & action/interaction & R1, R3, G2 \\
H2O-3D~\cite{hampali2022keypoint} & 2022 & RGB-D & multi-view & controlled & 3D joints (2H) & 6D pose & -- & R1, R3 \\
HOI4D~\cite{liu2022hoi4d} & 2022 & RGB-D & ego & controlled & MANO & 6D pose+state & action/interaction & R3, ET \\
OakInk~\cite{yang2022oakink} & 2022 & RGB & multi-view & controlled & MANO & shape+pose & intent/text & R1, G1 \\
ARCTIC~\cite{fan2023arctic} & 2023 & RGB & multi-view & controlled & MANO (2H) & art. pose+mesh & contact/action & R1, R3, G2 \\
AssemblyHands~\cite{ohkawa2023assemblyhands} & 2023 & RGB & ego & controlled & 3D joints & -- & -- & R1 (hand-only) \\
SHOWMe~\cite{swamy2023showme} & 2023 & RGB-D & exo & controlled & hand mesh & object mesh & -- & R2, R3 \\
AffordPose~\cite{jian2023affordpose} & 2023 & 3D assets & N/A & synthetic & MANO & shape+pose & affordance & G1 \\
GazeHOI~\cite{tian2024gaze} & 2024 & MoCap+gaze & N/A & controlled & hand motion & object motion & gaze/contact & G2 \\
HOGraspNet~\cite{cho2024dense} & 2024 & RGB-D & multi-view & controlled & MANO mesh/joints & object mesh & contact/grasp & R1, G1 \\
OakInk2~\cite{zhan2024oakink2} & 2024 & RGB & multi-view & controlled & MANO (2H) & shape+pose & intent/text & R1, R3, G2 \\
TACO~\cite{liu2024taco} & 2024 & RGB & ego+exo & controlled & hand motion & object motion & tool/action & G1, G2 \\
GigaHands~\cite{fu2025gigahands} & 2025 & RGB & multi-view & controlled & MANO (2H) & shape+pose & masks/text & R1, R3, G2 \\
HO-Cap~\cite{wang2026ho} & 2025 & RGB-D & ego+multi-view & controlled & MANO & 6D pose+mesh & tracks & R1, R3, ET \\
HOT~\cite{yu2025dynamic} & 2025 & RGB-D+tactile & multi-view & synthetic & 3D joints & deformable state & force/contact & R3 \\
HOT3D~\cite{banerjee2025hot3d} & 2025 & RGB & ego+multi-view & mixed & MANO (2H) & 6D pose & gaze/scene points & R1, R3 \\
ArtHOI-RGBD~\cite{wang2026arthoi} & 2026 & RGB-D & exo & controlled & -- & mesh+part motion & contact & R3 \\
EPIC-Contact~\cite{bansal2026hopformer} & 2026 & RGB & ego & in-the-wild & MANO (2H) & 6D pose+mesh & dense contact & R1 \\
DexGloveHOI~\cite{kou2026aviht} & 2026 & RGB+IMU & ego & controlled & 3D hand pose & -- & -- & R1 (hand-only) \\
SHOW3D~\cite{rim2026show3d} & 2026 & RGB & ego+exo & in-the-wild & 3D joints (2H) & 6D pose & captions & R1, R3 \\
HandX~\cite{zhang2026handx} & 2026 & MoCap & N/A & controlled & MANO (2H) & mesh+motion & contact/text & G2 \\
HUG-Bench~\cite{wu2026human} & 2026 & RGB-D & object & controlled & MANO & metric mesh & -- & G1, ET \\
\midrule
\multicolumn{9}{@{}l}{\textbf{B. Large-scale pretraining sources and embodied data engines}} \\
\midrule
Dataset & Year & Observation & View & Setting & \multicolumn{3}{c}{Supervision} & Use \\
\midrule
ActivityNet~\cite{DBLP:conf/cvpr/HeilbronEGN15} & 2015 & RGB & mixed & web & \multicolumn{3}{c}{action labels} & video-model pretraining \\
Charades~\cite{sigurdsson2016hollywood} & 2016 & RGB & exo & controlled & \multicolumn{3}{c}{action labels} & video-model pretraining \\
Kinetics~\cite{kay2017kinetics} & 2017 & RGB & mixed & web & \multicolumn{3}{c}{action labels} & video-model pretraining \\
Something-Something~\cite{DBLP:conf/iccv/GoyalKMMWKHFYMH17} & 2017 & RGB & exo & controlled & \multicolumn{3}{c}{action labels} & video-model pretraining \\
AVA~\cite{gu2018ava} & 2018 & RGB & exo & web & \multicolumn{3}{c}{spatiotemporal action labels} & video-model pretraining \\
HACS~\cite{zhao2019hacs} & 2019 & RGB & mixed & web & \multicolumn{3}{c}{action segments} & video-model pretraining \\
HowTo100M~\cite{miech2019howto100m} & 2019 & RGB & mixed & web & \multicolumn{3}{c}{narration/ASR} & video-model pretraining \\
FineGym~\cite{shao2020finegym} & 2020 & RGB & exo & web & \multicolumn{3}{c}{fine-grained actions} & video-model pretraining \\
EPIC-KITCHENS~\cite{damen2020epic} & 2021 & RGB & ego & in-the-wild & \multicolumn{3}{c}{narration/action labels} & video-model pretraining \\
Assembly101~\cite{sener2022assembly101} & 2022 & RGB & multi-view & controlled & \multicolumn{3}{c}{action/procedure labels} & video-model pretraining \\
Ego4D~\cite{grauman2022ego4d} & 2022 & RGB & ego & in-the-wild & \multicolumn{3}{c}{narration} & video-model pretraining \\
HoloAssist~\cite{wang2023holoassist} & 2023 & RGB-D+IMU & ego & controlled & \multicolumn{3}{c}{instruction/gaze/audio/hand pose} & video-model pretraining \\
Ego-Exo4D~\cite{grauman2024ego} & 2024 & RGB & ego+exo & mixed & \multicolumn{3}{c}{narration/body+hand pose} & video-model pretraining \\
EgoDex~\cite{hoque2025egodex} & 2025 & RGB & ego & in-the-wild & \multicolumn{3}{c}{language/camera/3D hand pose} & human-data policy pretraining \\
OpenEgo~\cite{jawaid2025openego} & 2025 & multimodal & ego & in-the-wild & \multicolumn{3}{c}{standardized signals/3D hand pose} & human-data policy pretraining \\
VITRA~\cite{li2025scalable} & 2025 & RGB & ego & web & \multicolumn{3}{c}{action chunks/hand motion} & human-data policy pretraining \\
EgoLive~\cite{li2026egolive} & 2026 & stereo RGB+IMU & ego & in-the-wild & \multicolumn{3}{c}{depth/3D pose/masks/tracks/language} & human-data policy pretraining \\
EgoScale~\cite{zheng2026egoscale} & 2026 & RGB & ego & web & \multicolumn{3}{c}{hand trajectories/camera/action labels} & human-data policy pretraining \\
EgoVerse~\cite{punamiya2026egoverse} & 2026 & RGB & ego & web & \multicolumn{3}{c}{camera/head motion/text} & human-data policy pretraining \\
HumanNet~\cite{deng2605humannet} & 2026 & RGB & ego+exo & web & \multicolumn{3}{c}{captions/motion} & video-model pretraining \\
FEEL~\cite{dessalene2026feel} & 2026 & RGB+force & ego & in-the-wild & \multicolumn{3}{c}{force/contact} & video-model pretraining \\
Open-AoE~\cite{li2026openaoe} & 2026 & RGB & ego & in-the-wild & \multicolumn{3}{c}{text/MANO/camera/action segments} & human-data policy pretraining \\
ACE-Data-0~\cite{cao2026acedata} & 2026 & multimodal RGB/video+motion+audio+tactile & ego+exo & in-the-wild & \multicolumn{3}{c}{body/hand/object trajectories; contact and language labels} & human-data policy pretraining \\
Ego2Robot~\cite{wang2026ego2robot} & 2026 & RGB video+motion & ego & in-the-wild & \multicolumn{3}{c}{robot actions/video; quality labels} & human-data policy pretraining \\
HandEdit~\cite{yang2026handedit} & 2026 & RGB (edit pairs) & ego & mixed & \multicolumn{3}{c}{human-to-robot edited pairs; 26 URDF embodiments} & human-data policy pretraining \\
RoboWheel~\cite{zhang2026robowheel} & 2026 & RGB & mixed & in-the-wild & \multicolumn{3}{c}{robot trajectories/retargeted demonstrations} & human-data policy pretraining \\
EgoEngine~\cite{liu2026egoengine} & 2026 & RGB & ego & in-the-wild & \multicolumn{3}{c}{hand-object trajectories/dexterous robot demonstrations} & human-data policy pretraining \\
EgoInfinity~\cite{wang2026egoinfinity} & 2026 & RGB & ego+exo & web & \multicolumn{3}{c}{4D HOI/robot-aligned action data} & human-data policy pretraining \\
Do as I Do~\cite{paliwal2026dexterous} & 2026 & RGB & ego & in-the-wild & \multicolumn{3}{c}{hand-object trajectories/robot demonstrations} & human-data policy pretraining \\
RoboEdit~\cite{liu2026roboedit} & 2026 & RGB & mixed & in-the-wild & \multicolumn{3}{c}{robotized visual experience} & human-data policy pretraining \\
Human2Robot~\cite{xie2026human2robot} & 2026 & RGB & mixed & mixed & \multicolumn{3}{c}{paired human-robot videos/actions} & human-data policy pretraining \\
\bottomrule
\end{tabular}%
}
\end{table*}

\subsection{Metric Families and Their Blind Spots}\label{sec:metrics}

\begin{table*}[!t]
\centering
\scriptsize
\setlength{\tabcolsep}{2pt}
\renewcommand{\arraystretch}{1.0}
\caption{Common metric families in HOI evaluation.}
\label{tab:metrics}
\begin{tabularx}{\textwidth}{@{}>{\raggedright\arraybackslash}p{0.21\textwidth}X>{\raggedright\arraybackslash}p{0.34\textwidth}@{}}
\toprule
Metric family & Common metrics & Applicable task(s) \\
\midrule
Hand pose/mesh accuracy & MPJPE, PA-MPJPE, MPVPE, PA-MPVPE, PCK & R1, R3; G1--G3 when ground-truth hand state exists \\
Object pose/shape accuracy & ADD/ADD-S, rotation/translation error, Chamfer distance, F-score, IoU & R1--R3; G2 when ground-truth object state exists \\
Contact agreement & Contact precision/recall/F1/IoU, contact distance & R1, R3; G1--G3 when contact annotations exist \\
Physical plausibility & Penetration depth/volume, simulation displacement, force closure & R1, R3, G1, G2 \\
Image/video distribution fidelity & FID, FVD & G3 \\
Paired visual similarity & PSNR, SSIM, LPIPS & R2/R3 rendered outputs; paired G3 editing/reenactment \\
Generative diversity/coverage & Diversity, coverage/multimodality, condition-adherence rate & G1--G3, when multiple samples or conditions are evaluated \\
Motion/trajectory accuracy & Sequence MPJPE/MPVPE, hand/object trajectory error & R3, G2 \\
Embodied-transfer evaluation & Retargeting error, task success rate, object-state success & ET \\
\bottomrule
\end{tabularx}
\end{table*}

\subsubsection{Geometry Metrics}\label{sec:metrics-geometry}
Joint and vertex errors such as MPJPE and MPVPE require paired 3D ground truth and measure hand-pose or mesh accuracy in the evaluation coordinate frame. Their Procrustes-aligned variants, PA-MPJPE and PA-MPVPE, first align prediction and ground truth through a similarity transformation that removes global translation, rotation, and scale, emphasizing articulated pose/shape agreement while concealing errors in absolute position, orientation, and metric scale~\cite{DBLP:journals/pami/IonescuPOS14}. Object metrics have distinct data requirements: rotation and translation errors require ground-truth object pose; ADD and ADD-S additionally require object model points; and surface metrics such as Chamfer distance, F-score, and IoU require ground-truth surface or occupancy geometry, but not necessarily a posed CAD model~\cite{hinterstoisser2012model,xiang2017posecnn,fan2017point}. These metrics quantify geometric agreement, but lower error does not establish correct contact, graspability, or functional interaction.

\subsubsection{Visual and Motion Metrics}\label{sec:metrics-visual}

FID and FVD compare unpaired image and video distributions, respectively, whereas PSNR, SSIM, and LPIPS require paired references and are better suited to reconstruction, editing, or reenactment~\cite{DBLP:conf/nips/HeuselRUNH17,unterthiner2018towards,wang2004image,zhang2018unreasonable}. Motion and trajectory errors also require paired sequences, while diversity and coverage characterize the spread of generated samples rather than their correctness. For conditional generation, a condition-adherence rate measures whether outputs satisfy the requested text, pose, contact, or motion condition, but does not establish geometric/physical validity. None of these measures verifies hand anatomy, 3D consistency, contact persistence, task completion, or physical dynamics.

\subsubsection{Contact and Physical Metrics}\label{sec:metrics-contact}

Contact precision, recall, F1, IoU, and distance require contact annotations or registered hand-object surfaces and measure spatial overlap between predicted and ground-truth contact regions~\cite{hasson2019learning,jiang2021hand}. Penetration, simulation displacement, and force-closure measures instead test complementary aspects of physical plausibility but depend strongly on mesh quality, friction, contact modeling, and simulators. High contact prediction accuracy does not guarantee force closure, execution stability, or functional correctness, and a favorable physical score does not by itself establish functional task success.

Embodied-transfer evaluation spans two complementary levels. Retargeting error measures the fidelity of mapping a human-derived grasp or trajectory to a robot embodiment, whereas task success rate and object-state success test whether the transferred signal supports the intended robot behavior. Mapping accuracy is therefore informative but should not be conflated with downstream task performance.


\section{Open Challenges and Future Directions}\label{sec:challenges}


\subsection{Toward Integrated and Verifiable HOI Foundation Systems}\label{sec:unified-architecture}

No single prior family resolves HOI: geometric priors improve shape and alignment, semantic priors localize function and intent, and visual priors model appearance and dynamics, yet each can fail. Integrated systems should compose them around shared hand-object states, contacts, and trajectories according to the dominant uncertainty, then verify contact, physical plausibility, and, where applicable, robot execution.

\subsection{From Geometric Correctness to Interaction Correctness}\label{sec:interaction-correctness}

HOI evaluation remains dominated by MPJPE, Chamfer distance, and ADD. Although ContactDB~\cite{brahmbhatt2019contactdb}, ContactPose~\cite{brahmbhatt2020contactpose}, GRAB~\cite{taheri2020grab}, ARCTIC~\cite{fan2023arctic}, and EPIC-Contact~\cite{bansal2026hopformer} provide contact annotations, protocols rarely assess contact together with function, physical plausibility, and object-state transitions. Future benchmarks should unify these dimensions and distinguish proximity from contact that transmits force and produces the intended response.

\subsection{Long-Horizon, Dynamic-Camera, and World-Space HOI}\label{sec:long-horizon}

Egocentric HOI entangles camera and hand-object motion across extended manipulation, causing camera-frame estimates to drift. HOSt3R~\cite{swamy2025host3r} replaces SfM with DUSt3R~\cite{wang2024dust3r} pointmaps, EgoGrasp~\cite{fu2026egograsp} targets world-space HOI, and ArtHOI~\cite{wang2026arthoi} handles articulated-object 4D reconstruction. Jointly recovering ego-motion, contact, and long-horizon hand-object state in one world frame across grasp, manipulation, release, and re-grasp remains open.

\subsection{Prior Reliability, Routing, and Conflict Resolution}\label{sec:prior-routing}

Combined priors may disagree: retrieval can return a mismatched asset, reconstruction can hallucinate contact-inconsistent geometry, grounding can select the wrong functional part, and generated video can hide invalid physics. Systems should estimate confidence, route priors using visibility, camera motion, occlusion, and intent, and expose conflicts rather than average incompatible evidence.

\subsection{From HOI Analysis to Dynamic Embodied Memory}\label{sec:pregrasp-anchoring}

After grasping, the gripper often occludes the evolving object relationship. Pre-contact reconstruction and grounding can capture object identity, geometry, pose, functional regions, and intended contact before this evidence disappears. A dynamic embodied memory should preserve these variables and their uncertainty, update them with vision, proprioception, contact events, and object responses, and maintain contact history and state transitions across grasp, manipulation, release, and re-grasp. The challenge is reconciling pre-grasp estimates with later evidence under self-occlusion and changing viewpoints.

\subsection{Robot-Centric Understanding of Human HOI}\label{sec:robot-centric-understanding}

Embodied transfer often converts human HOI into kinematic trajectories without intent, object/part identity, contact, and the resulting scene state change~\cite{ye2025latent,wang2025gat,DBLP:conf/rss/WenLS0D0A24,chen2025vidbot}. Robust cross-embodiment transfer requires robot-usable descriptions that include these variables and the interaction phase, enabling policies to replan under perturbations rather than replay memorized motion. Evaluation should therefore measure intent, contact timing, object-state outcomes, and safety during execution rather than trajectory similarity alone.


\section{Conclusion}\label{sec:conclusion}

This survey examined HOI reconstruction and generation through the lens of foundation-model priors. We organized existing methods into geometric, semantic, and visual prior families and analyzed how their knowledge enters HOI pipelines. We further reviewed how HOI-derived signals support embodied transfer. Despite rapid progress, challenges remain in composing heterogeneous priors, ensuring interaction and physical correctness, and reasoning over long-horizon open-world interactions. We hope this survey provides guidance for understanding current progress and developing generalizable, interaction-aware, and physically grounded HOI systems.

\bibliographystyle{unsrtnat}
\bibliography{references}

\end{document}